\documentclass[10pt,twocolumn,letterpaper]{article}

\usepackage[pagenumbers]{cvpr}
\usepackage{url}
\usepackage{graphicx}

\usepackage{booktabs}       
\usepackage{amsfonts}       
\usepackage{nicefrac}       
\usepackage{xcolor}         

\usepackage{amsmath}
\usepackage{amssymb}
\usepackage{graphicx}
\usepackage{multirow}
\usepackage{bm}
\usepackage{array}
\usepackage{makecell}
\usepackage{adjustbox}
\usepackage[table]{xcolor}
\usepackage{pifont}
\usepackage{algorithm}
\usepackage{algpseudocode}
\usepackage{wrapfig}
\usepackage{cuted}
\usepackage{capt-of}

\newcommand{\sd}[1]{{\tiny$\pm$#1}}

\definecolor{cvprblue}{rgb}{0.21,0.49,0.74}
\usepackage[pagebackref,breaklinks,colorlinks,allcolors=cvprblue]{hyperref}

\title{MoSk: A Unified Motion-Skeleton Foundation Model\\for Human Motion Understanding and Generation}

\author{Yun Chen\\
[0.3em]
KAIST\\
{\tt\small cyruby@kaist.ac.kr}
\and
Munchurl Kim\footnotemark[2]\\
[0.3em]
KAIST\\
{\tt\small mkimee@kaist.ac.kr}
\and
Jeonghyeok Do\footnotemark[2]\\
[0.3em]
KAIST\\
{\tt\small ehwjdgur0913@kaist.ac.kr}
\and
\small{\url{https://kaist-viclab.github.io/MoSk_site}}
}

\begin{document}
\maketitle

{
  \renewcommand{\thefootnote}%
    {\fnsymbol{footnote}}
  \footnotetext[2]{Co-corresponding authors (equal advising).}
}

\begin{abstract}
Foundation models (FMs) have advanced visual understanding and generation by providing semantically rich representations. However, this role of FMs remains underexplored in human motion. Large motion datasets offer diverse movements at scale, but often lack action labels for systematically evaluating learned representations on understanding tasks. Skeleton datasets enable such evaluation through established recognition protocols, yet are typically smaller and focus on predefined actions captured in controlled environments. To combine these strengths, we present \textbf{MoSk}, to the best of our knowledge, the first \textbf{Mo}tion-\textbf{Sk}eleton foundation model to integrate motion and skeleton datasets. MoSk learns shared representations through self-supervised masked motion modeling. We incorporate joint-name embeddings to provide a common anatomical reference across various skeleton formats. A cross-attention decoder reconstructs masked motion from the encoder's features, with joint-name embeddings specifying the target joints. Experiments demonstrate strong action-recognition performance across diverse settings, including linear evaluation and fine-tuning. MoSk also substantially improves motion generation across multiple backbones through representation alignment, reducing FID on MotionMillion by up to 9.75 for unconditional generation and 3.83 for text-to-motion generation.
\end{abstract}

\section{Introduction}
\label{sec:introduction}

\begin{figure}[t]
    \centering
    \includegraphics[width=\linewidth]{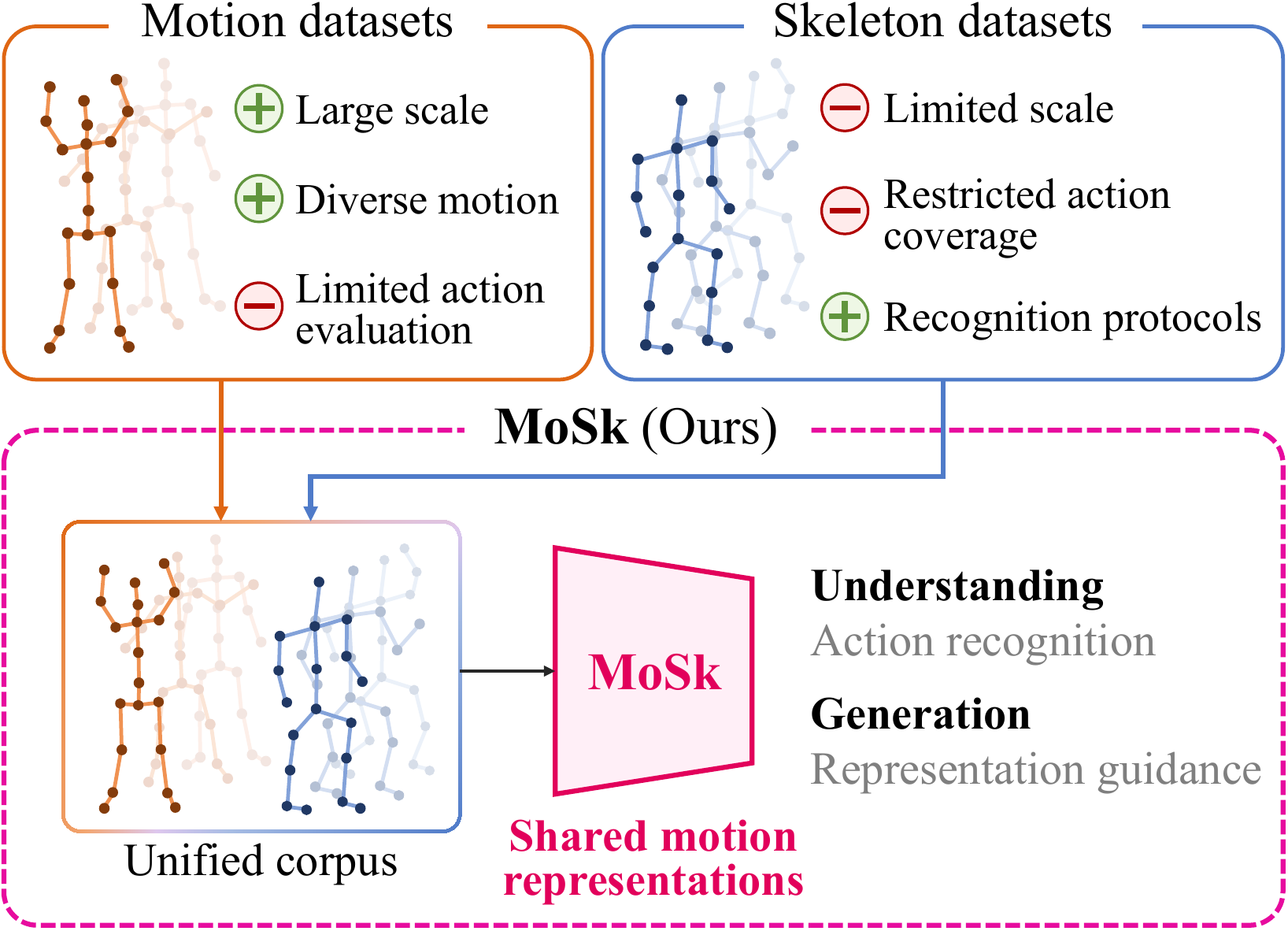}
    \vspace{-0.6cm}
    \caption{\textbf{Motivation.} Motion datasets are large and diverse but often lack recognition protocols. Skeleton datasets provide such protocols but have limited scale and action coverage. MoSk integrates both sources into a unified corpus and is trained as a motion foundation model for both understanding and generation.}
    \label{fig:motivation}
    \vspace{-0.3cm}
\end{figure}


Foundation models (FMs) are trained on large, diverse datasets to learn general representations, providing a common basis for multiple downstream tasks. These representations have been widely used for understanding tasks, including classification, segmentation, and object detection~\cite{he2022mae,oquab2023dinov2}. More recently, representations learned by foundation models have also been used to guide generative training through representation alignment and to provide latent spaces for generation~\cite{yu2025repa,zheng2026rae}. These developments connect advances in visual understanding with generative modeling, enabling faster convergence and improved sample quality. A key approach to building such models is self-supervised learning (SSL), which uses the data itself as supervision without task-specific labels~\cite{he2022mae,caron2021dino,oquab2023dinov2}. This reduces dependence on manual annotation and enables learning from broader data sources. However, learning human motion representations that transfer broadly across understanding and generation remains an open challenge.

Motion datasets for generation and skeleton datasets for action recognition are two major sources of human motion data. Motion datasets commonly pair motion sequences with text captions for text-to-motion generation~\cite{guo2022humanml3d,plappert2016kit}, while recent large-scale datasets further expand the diversity and coverage of human movements~\cite{zhang2026romo,fan2025motionmillion}. However, their evaluation remains centered on generation, with limited categorical annotations and benchmark protocols for evaluating action understanding. Skeleton datasets, in contrast, provide action labels and established evaluation protocols~\cite{shahroudy2016ntu,liu2019ntu120,liu2017pku}, but their limited sizes, fixed action sets, and constrained recording settings restrict the breadth of motion available for training. These complementary sources give a single motion-skeleton foundation model access to a richer range of movements for training, together with labeled benchmarks that assess its understanding capabilities. Fig.~\ref{fig:motivation} summarizes the complementary strengths of motion and skeleton datasets and motivates bringing them together in a unified training corpus.

Combining the strengths of motion data and skeleton data, we introduce \textbf{MoSk}, a unified \textbf{mo}tion-\textbf{sk}eleton foundation model for human motion understanding and generation. To the best of our knowledge, MoSk is the \textit{first motion-skeleton foundation model} to learn shared representations from both motion and skeleton datasets. We standardize heterogeneous sequences into a common format of 3D joint positions over time, integrating both sources into a single training corpus. A single motion encoder is trained on this corpus through self-supervised masked modeling~\cite{he2022mae}.

Within this common format of 3D joint positions, datasets retain their original joint sets, which differ in joint counts and anatomical definitions. Inspired by SOfA~\cite{do2026sofa}, we incorporate joint-name embeddings into motion encoding to provide a shared anatomical reference across skeleton formats. In masked motion modeling, parts of each sequence are masked, and the encoder extracts features from the remaining motion. A cross-attention decoder then uses these features to reconstruct the masked motion, with joint-name embeddings specifying the target joints. This masked modeling objective encourages the encoder to learn semantically meaningful motion representations that capture spatial dependencies among joints and how their movements change over time. The motion encoder trained through this process serves as MoSk, providing a common foundation of motion knowledge for both understanding and generation.

We conduct extensive experiments on both high-level action understanding and low-level motion generation to examine MoSk's transferability. For skeleton-based action recognition, a classification head is trained on representations extracted by MoSk. For motion generation, we align the features of different generative backbones with MoSk representations during training~\cite{yu2025repa}. Each backbone is compared with its counterpart trained without MoSk to assess the benefit of representation guidance across model architectures. MoSk's representations are evaluated as direct inputs for action recognition and as training supervision for motion synthesis. MoSk outperforms dataset-specific specialists on multiple benchmarks under linear evaluation and fine-tuning, and also transfers to unseen skeleton formats. In linear evaluation, it reaches 94.0\% accuracy on NTU RGB+D 60 (X-View) and 69.4\% on PKU-MMD~II. MoSk guidance reduces FID in 15 of 16 settings across four datasets and two backbones, by up to $5.67$ (unconditional) and $2.08$ (text-to-motion) on RoMo.

Our contributions are threefold:
\begin{itemize}
\item We introduce \textbf{MoSk}, the \textit{first motion-skeleton foundation model} that learns shared representations from motion and skeleton data through self-supervised training, and provides a single trained encoder for both motion understanding and generation.
\item Motion and skeleton datasets are integrated into a unified training corpus, increasing the scale and diversity of data available for human motion representation learning.
\item Extensive experiments demonstrate that MoSk outperforms dataset-specific specialists on multiple action-recognition benchmarks and substantially improves motion generation across different generative backbones.
\end{itemize}


\section{Related Work}
\label{sec:related_work}

\begin{figure*}[t]
    \centering
    \includegraphics[width=1.0\linewidth]{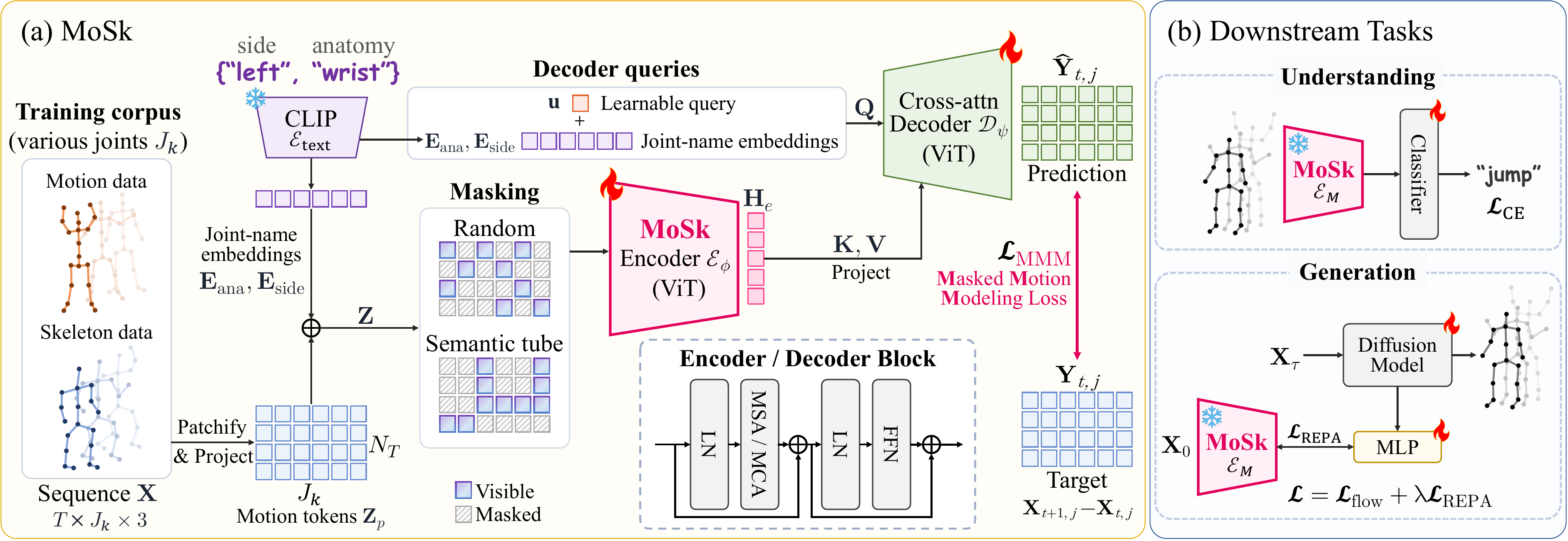}
    \vspace{-0.6cm}
    \caption{\textbf{Overview of MoSk.} (a) MoSk learns from a combined corpus of motion and skeleton sequences with different joint sets through masked modeling. Joint-name embeddings provide a shared anatomical reference across these formats, while masked modeling encourages the encoder to capture spatial and temporal motion structure. Decoder queries combine a shared learnable vector with joint-name embeddings to specify which joints to reconstruct from the encoded context. Each encoder/decoder block contains attention (MSA/MCA) and feed-forward (FFN) layers, each preceded by layer normalization (LN) and paired with a residual connection. (b) MoSk's frozen encoder (except during semi-supervised and full fine-tuning) supports action recognition via a classification head and guides motion generators via training-time feature alignment. Snowflakes and flames denote frozen and trainable modules, respectively.}
    \label{fig:framework}
    \vspace{-0.3cm}
\end{figure*}

\subsection{Visual Representation Learning}
\label{sec:related_visual}

\noindent\textbf{Self-supervised representation learning.}
Self-supervised learning (SSL) derives supervision from the data itself, enabling representation learning without task-specific labels. Contrastive learning learns visual representations by bringing augmented views of the same image closer while distinguishing different images~\cite{chen2020simclr,he2020momentum}. Masked autoencoding provides another learning objective, reconstructing missing regions in images or videos from partial observations~\cite{he2022mae,tong2022videomae}. Self-distillation approaches learn transferable visual features by matching teacher and student outputs across different views, with large-scale training further improving their generalization~\cite{caron2021dino,oquab2023dinov2,simeoni2025dinov3}. Feature prediction approaches learn semantic and temporal structures by predicting representations of missing image regions or video segments from visible context~\cite{assran2023ijepa,bardes2024vjepa}. Together, these approaches provide a foundation for learning general visual representations without task-specific annotations.

\noindent\textbf{Visual representations for generation.}
Pretrained visual representations support generative modeling through two main directions: 
(i) alignment-based approaches use pretrained features to supervise generative features, as in Representation Alignment (REPA)~\cite{yu2025repa} and REPA-E~\cite{leng2025repae}, or tokenizer latents, as in VA-VAE~\cite{yao2025vavae}, improving training efficiency and generation quality; (ii) another direction directly incorporates pretrained representation encoders into autoencoding frameworks and trains generative models in the resulting latent spaces~\cite{zheng2026rae,bi2026vfmvae}. These studies demonstrate the usefulness of visual representations both as training supervision and as generative latent spaces.

\subsection{Human Motion Representation Learning}
\label{sec:related_motion}

\noindent\textbf{Skeleton representations for action understanding.}
Self-supervised skeleton representation learning captures spatial and temporal motion structures for action understanding. Contrastive learning approaches encourage consistent representations across augmented sequences and different skeletal views~\cite{li2021crossclr,lin2023actionlet}. Masked prediction approaches learn from incomplete observations by recovering joint motion or predicting latent features~\cite{mao2023mamp,abdelfattah2024sjepa,do2026slim}. Broader pretraining frameworks further explore representations that transfer across multiple human motion understanding tasks~\cite{zhu2023motionbert,weng2025usdrl}.

\noindent\textbf{Motion representations for generation.}
Motion generation methods learn motion representations to support text-conditioned synthesis~\cite{chen2023mld,zhang2023t2mgpt,guo2024momask,li2026motionhiflow}. Another line of work connects motion representations with language through cross-modal alignment and paired pretraining, improving semantic consistency in text-conditioned motion generation~\cite{tevet2022motionclip,li2025lamp}. Our MoSk learns representations from both skeleton action and motion-generation datasets through self-supervised learning, without action labels or motion captions. These representations support both action understanding and motion generation.

\section{Method}
\label{sec:method}

MoSk is a motion-skeleton foundation model that learns transferable representations from heterogeneous motion and skeleton datasets through self-supervised learning. Fig.~\ref{fig:framework} provides an overview of MoSk. First, input sequences from various motion and skeleton datasets are standardized to a common geometric and temporal format while retaining each dataset's joint set (Sec.~\ref{sec:unifying_data}). The encoder learns motion structure through masked motion modeling with joint-name embeddings in input tokens and a learnable decoder query (Sec.~\ref{sec:ssl_motion}). The trained encoder provides representations for action recognition and feature-level supervision for motion generation (Sec.~\ref{sec:downstream}).

\subsection{Unifying Skeleton and Motion Data}
\label{sec:unifying_data}

We combine skeleton and motion datasets into a unified training corpus. Training uses only the input sequences, without action labels or motion captions. For an input sequence $S_k$ from the $k$-th dataset, the corresponding preprocessing operator $\mathcal{A}_k$ converts $S_k$ from its original format into a common 3D joint-coordinate format:

\begin{equation}
    \mathbf{X}=\mathcal{A}_k(S_k)
    \in\mathbb{R}^{T\times J_k\times C_{\mathrm{in}}},
    \qquad C_{\mathrm{in}}=3,
    \label{eq:canonical_data}
\end{equation}

\noindent where $J_k$ denotes the joint count of the $k$-th dataset. Specifically, $\mathcal{A}_k$ normalizes the initial translation, heading, and scale in a common coordinate system with the $y$-axis pointing upward, the ground plane at $y=0$, and the initial facing direction along $+z$. It also samples $T=64$ frames from the input sequence $S_k$. This preprocessing standardizes the coordinate system and sequence length while preserving each dataset's original joint set, allowing skeleton and motion datasets to be used together for training.

\subsection{Self-Supervised Motion Learning}
\label{sec:ssl_motion}

\noindent\textbf{Motion encoder.}
Fig.~\ref{fig:framework} shows the motion encoder $\mathcal{E}_{\phi}$, which uses a Vision Transformer (ViT) backbone. Each input sequence $\mathbf{X}$ is partitioned into non-overlapping temporal segments of $P_T=4$ consecutive frames, with each joint in a segment forming one patch. These patches are reshaped into $\mathbf{X}_p\in\mathbb{R}^{N_T\times J_k\times C_p}$, where $N_T=T/P_T$ is the number of temporal patches per joint and $C_p=P_TC_{\mathrm{in}}$ is the flattened patch dimension. A shared linear projection $\mathbf{W}_x\in\mathbb{R}^{C_p\times C_e}$ embeds each patch into $C_e$ dimensions as:
\begin{equation}
    \mathbf{Z}_p=\mathbf{X}_p\mathbf{W}_x
    \in\mathbb{R}^{N_T\times J_k\times C_e}.
    \label{eq:patch_embedding}
\end{equation}

Since skeleton formats differ in their joint definitions and numbering across datasets, we address these differences through a unified naming scheme that describes each joint $j$ using two text strings: its anatomical name $\mathbf{t}_j^{\mathrm{ana}}$, such as \emph{wrist} or \emph{knee}, and its side $\mathbf{t}_j^{\mathrm{side}}\in\{\mathrm{left},\mathrm{right},\mathrm{middle}\}$. A frozen CLIP~\cite{radford2021clip} text encoder $\mathcal{E}_{\mathrm{text}}$ encodes these strings separately to produce joint-name embeddings as:
\begin{equation}
(\mathbf{E}_{\mathrm{ana}})_{j}=\mathcal{E}_{\mathrm{text}}(\mathbf{t}_j^{\mathrm{ana}}),\quad (\mathbf{E}_{\mathrm{side}})_{j}=\mathcal{E}_{\mathrm{text}}(\mathbf{t}_j^{\mathrm{side}}),
\label{eq:joint_name_embeddings}
\end{equation}
where $\mathbf{E}_{\mathrm{ana}},\mathbf{E}_{\mathrm{side}}\in\mathbb{R}^{J_k\times C_{\mathrm{text}}}$ and $C_{\mathrm{text}}$ is the text embedding dimension. The two embeddings are mapped to the encoder dimension by trainable projections $\mathbf{W}^{e}_{\mathrm{ana}},\mathbf{W}^{e}_{\mathrm{side}}\in\mathbb{R}^{C_{\mathrm{text}}\times C_e}$, respectively, and added to the motion tokens $\mathbf{Z}_p$ to form the encoder input embeddings $\mathbf{Z}$ as:
\begin{equation}
    \mathbf{Z}=\mathbf{Z}_p+\mathbf{E}_{\mathrm{ana}}\mathbf{W}^{e}_{\mathrm{ana}}+\mathbf{E}_{\mathrm{side}}\mathbf{W}^{e}_{\mathrm{side}}.
    \label{eq:input_embedding}
\end{equation}

\noindent The projected joint-name embeddings are repeated along the temporal dimension, giving $\mathbf{Z}\in\mathbb{R}^{N_T\times J_k\times C_e}$. These embeddings provide a shared anatomical reference, enabling the encoder $\mathcal{E}_{\phi}$ to handle different skeleton formats.

To train MoSk through masked motion modeling, we initially drop a random subset of joints from the input across all frames. Token masking is then applied using semantic tube masking from SLiM~\cite{do2026slim} or uniform random masking with equal probability. Semantic tube masking masks connected anatomical regions over consecutive temporal positions, whereas  random masking selects individual tokens.

Let $\mathcal{V}$ contain the joint-temporal indices of the tokens that remain visible on the $N_T\times J_k$ token grid. The unmasked tokens are gathered and flattened as:
\begin{equation}
\begin{aligned}
    \mathbf{Z}^{u}
    &=\operatorname{Gather}(\mathbf{Z},\mathcal{V})
      \in\mathbb{R}^{N_u\times C_e},\\
    N_u&=|\mathcal{V}|=N_TJ_k(1-r),
\end{aligned}
\label{eq:visible_tokens}
\end{equation}
where $N_u$ is the number of visible tokens and $r$ is the total masking ratio. The encoder $\mathcal{E}_{\phi}$ processes the visible tokens through $L_e$ Transformer encoder blocks to obtain the encoded features $\mathbf{H}_e$ as:
\begin{equation}
\begin{aligned}
    \mathbf{H}_0
    &=\mathbf{Z}^{u},\\
    \mathbf{H}'_l
    &=\operatorname{MSA}
      (\operatorname{LN}(\mathbf{H}_{l-1}))
      +\mathbf{H}_{l-1},\\
    \mathbf{H}_l
    &=\operatorname{FFN}
      (\operatorname{LN}(\mathbf{H}'_l))
      +\mathbf{H}'_l,\\
    \mathbf{H}_e
    &=\operatorname{LN}(\mathbf{H}_{L_e})
      \in\mathbb{R}^{N_u\times C_e},
\end{aligned}
\label{eq:mosk_encoder}
\end{equation}
for $l=1,\ldots,L_e$. Here, $\operatorname{MSA}$ denotes multi-head self-attention with 1D temporal rotary positional embeddings (RoPE). RoPE uses each token's original temporal patch index after masking and flattening. $\operatorname{FFN}$ is a SwiGLU feed-forward layer, and $\operatorname{LN}$ denotes layer normalization.

\noindent\textbf{Motion decoder.}
The ViT-based decoder $\mathcal{D}_{\psi}$ uses the same anatomy and side embeddings ($\mathbf{E}_\mathrm{ana}, \mathbf{E}_\mathrm{side}$) to specify which joints to reconstruct. The decoder queries $\mathbf{Q}$ are constructed from a shared learnable query vector $\mathbf{u}\in\mathbb{R}^{C_d}$ and the joint-name embeddings according to:
\begin{equation}
    \mathbf{Q}=\{\mathbf{u}\}+\mathbf{E}_{\mathrm{ana}}\mathbf{W}^{d}_{\mathrm{ana}}+\mathbf{E}_{\mathrm{side}}\mathbf{W}^{d}_{\mathrm{side}}\in\mathbb{R}^{N_T\times J_k\times C_d},
    \label{eq:decoder_query}
\end{equation}
where $\mathbf{W}^{d}_{\mathrm{ana}},\mathbf{W}^{d}_{\mathrm{side}}\in\mathbb{R}^{C_{\mathrm{text}}\times C_d}$ are trainable projections, and $\{\mathbf{u}\}$ denotes repeating the same query vector across joints and time while the projected joint-name embeddings are repeated only across time.
The queries $\mathbf{Q}$ are flattened into $\mathbf{Q}_0\in\mathbb{R}^{N_TJ_k\times C_d}$ and processed by $L_d$ Transformer decoder blocks. Each block uses the encoder output $\mathbf{H}_e$ as keys and values in its cross-attention layer as:
\begin{equation}
\begin{aligned}
    \mathbf{Q}'_l
    &=
    \operatorname{MCA}\!\left(
    \operatorname{LN}(\mathbf{Q}_{l-1}),
    \operatorname{LN}(\mathbf{H}_e)\right)
    +\mathbf{Q}_{l-1},\\
    \mathbf{Q}_l
    &=\operatorname{FFN}
      (\operatorname{LN}(\mathbf{Q}'_l))
      +\mathbf{Q}'_l,\\
    \mathbf{Q}_d
    &=\operatorname{LN}(\mathbf{Q}_{L_d})
      \in\mathbb{R}^{N_TJ_k\times C_d},
\end{aligned}
\label{eq:mosk_decoder}
\end{equation}
for $l=1,\ldots,L_d$, where $\operatorname{MCA}$ denotes multi-head cross-attention with 1D temporal RoPE using the temporal patch indices of queries and keys.

A linear prediction head $\mathbf{W}_o\in\mathbb{R}^{C_d\times C_p}$ maps the decoded features $\mathbf{Q}_d$ to motion patches according to:
\begin{equation}
    \widehat{\mathbf{Y}}_p=\mathbf{Q}_d\mathbf{W}_o
    \in\mathbb{R}^{N_TJ_k\times C_p}.
    \label{eq:motion_prediction}
\end{equation}
Reversing the patchification yields the framewise prediction $\widehat{\mathbf{Y}}\in\mathbb{R}^{T\times J_k\times C_{\mathrm{in}}}$.

\noindent\textbf{Training objective.}
Following MAMP~\cite{mao2023mamp}, we reconstruct joint velocities rather than the original joint coordinates. The target is the frame-to-frame displacement as:
\begin{equation}
    \mathbf{Y}_{t,j}
    =\mathbf{X}_{t+1,j}-\mathbf{X}_{t,j},
    \qquad 1\leq t<T.
    \label{eq:motion_target}
\end{equation}
We adopt the motion-energy-based focal weighting of AMR~\cite{sun2026amr} as:
\begin{equation}
    \mathcal{L}_{\mathrm{MMM}}
    =\frac{1}{|\Omega|}
      \sum_{(t,j)\in\Omega}
      \omega_{t,j}
      \big\|\widehat{\mathbf{Y}}_{t,j}
      -\mathbf{Y}_{t,j}\big\|_2^2,
    \label{eq:mosk_mmm}
\end{equation}
where $\Omega$ contains the valid joint-frame positions masked in the encoder input and $|\Omega|$ is the number of the positions, so the loss is computed only over the masked regions. The weights $\omega_{t,j}$ are computed from the input motion using AMR's focal weighting scheme, reducing the contribution of regions with low motion energy. The encoder $\mathcal{E}_{\phi}$, decoder $\mathcal{D}_{\psi}$, and embedding projections $\mathbf{W}^{e}_{\mathrm{ana}},\mathbf{W}^{e}_{\mathrm{side}}$,$\mathbf{W}^{d}_{\mathrm{ana}}$,$\mathbf{W}^{d}_{\mathrm{side}}$ are optimized jointly, while the text encoder $\mathcal{E}_{\mathrm{text}}$ remains frozen. 


\subsection{Understanding and Generation}
\label{sec:downstream}

MoSk ($\mathcal{E}_{M}$) consists of the trained encoder $\mathcal{E}_{\phi}$ and its input embedding module, providing the integrated motion and skeleton representations for understanding and generation.

\noindent\textbf{Action understanding.}
For skeleton-based action recognition, we convert an input sequence $S_k$ from the $k$-th dataset into $\mathbf{X}=\mathcal{A}_k(S_k)$ and predict action scores $\widehat{\mathbf{y}}$ using a linear classification head as:
\begin{equation}
    \widehat{\mathbf{y}}
    =\operatorname{Linear}\!\left(
    \operatorname{sg}[\mathcal{E}_{M}(\mathbf{X})]
    \right),
    \label{eq:action_prediction}
\end{equation}
where $\operatorname{sg}$ denotes stop-gradient and is omitted during semi-supervised and full fine-tuning. For action label $y$, the classification head is trained with cross-entropy loss as:
\begin{equation}
    \mathcal{L}_{\mathrm{CE}}
    =-\log\left[
    \operatorname{softmax}(\widehat{\mathbf{y}})
    \right]_{y}.
    \label{eq:action_loss}
\end{equation}
The classification head is trained using the target benchmark's action labels, and the encoder remains frozen during linear evaluation.

\noindent\textbf{Motion generation.}
For motion generation, we use MoSk as a frozen representation teacher following REPA's representation alignment approach~\cite{yu2025repa}. This approach uses representations of clean motion to supervise the generative model's intermediate features during denoising. Given a clean motion sequence $S_0$ from the $k$-th dataset, we obtain $\mathbf{X}_0=\mathcal{A}_k(S_0)$ and extract teacher features as:
\begin{equation}
    \mathbf{F}
    =\operatorname{sg}[
    \mathcal{E}_{M}(\mathbf{X}_0)].
    \label{eq:teacher_features}
\end{equation}

\noindent Let $\mathbf{X}_{\tau}$ be the noisy motion at generative time ${\tau}$ and $c$ denote the conditioning information, with $c=\varnothing$ for unconditional generation. We align the backbone's intermediate features $h_{\theta}(\mathbf{X}_{\tau},{\tau};c)$ with $\mathbf{F}$ through a trainable projection by minimizing the alignment loss as:
\begin{equation}
    \mathcal{L}_{\mathrm{REPA}}
    =\mathbb{E}\!\left[
    \ell_{\mathrm{align}}\!\left(
    \mathrm{MLP}(h_{\theta}(\mathbf{X}_{\tau},{\tau};c)),
    \mathbf{F}\right)\right],
    \label{eq:generative_alignment}
\end{equation}
where $\ell_{\mathrm{align}}$ denotes the feature alignment loss. The overall training objective is
\begin{equation}
    \mathcal{L}
    =\mathcal{L}_{\mathrm{flow}}
    +\lambda\mathcal{L}_{\mathrm{REPA}},
    \label{eq:generative_loss}
\end{equation}
where $\mathcal{L}_{\mathrm{flow}}$ is the backbone's original generative loss and the weighting hyperparameter is empirically set to $\lambda=0.5$. MoSk provides feature-level guidance only during training, introducing no additional inference cost.


\begin{table}[tbp]
    \scriptsize
    \centering
    \caption{\textbf{Training corpus.} Motion and skeleton datasets retain their native skeleton formats. Annotations are not used for training. Kinect~v1 and OpenNI formats are excluded from training for unseen-format evaluation.}
    \label{tab:data}
    \vspace{-0.2cm}
    \setlength{\tabcolsep}{3pt}
    \renewcommand{\arraystretch}{1.1}
    \resizebox{0.9\columnwidth}{!}{%
    \begin{tabular}{lrrr}
        \toprule
        \textbf{Dataset} & \textbf{Annotation} & \textbf{Clips} & \textbf{Hours} \\
        \midrule
        \rowcolor{gray!10}
        \multicolumn{4}{l}{\textit{Motion (SMPL, 22 joints / KIT, 21 joints)}} \\
        HumanML3D~\cite{guo2022humanml3d} & text & 12,326 & 24.2 \\
        KIT-ML~\cite{plappert2016kit} & text & 3,386 & 10.4 \\
        MotionMillion~\cite{fan2025motionmillion} & text & 595,422 & 819.6 \\
        RoMo~\cite{zhang2026romo} & text & 766,864 & 1,166.2 \\
        \rowcolor{gray!10}
        \multicolumn{4}{l}{\textit{Skeleton (Kinect v2, 25 joints)}} \\
        NTU RGB+D 60 / 120~\cite{shahroudy2016ntu,liu2019ntu120} & 120 actions & 18,757 & 12.8 \\
        PKU-MMD II~\cite{liu2017pku} & 41 actions & 728 & 5.7 \\
        ETRI-Activity3D~\cite{jang2020etri} & 55 actions & 67,622 & 136.4 \\
        \midrule
        Total (3 formats) & & 1,465,105 & 2,175.4 \\
        \bottomrule
    \end{tabular}}
    \vspace{-0.3cm}
\end{table}

\section{Experiments}
\label{sec:experiment}

\begin{table*}[t]
\centering
\caption{\textbf{Linear and semi-supervised evaluation} (top-1 accuracy, \%; joint stream). Linear evaluation freezes MoSk and uses all training clips. Semi-supervised evaluation fine-tunes MoSk and the classifier using uniformly sampled 1\% or 10\% subsets of the NTU-60 training splits. SOfA uses a filtered X-View test subset. \textbf{Bold}/\underline{underline} mark best/second-best results, excluding single-source MoSk variants.}
\label{tab:linear_semi}
\vspace{-0.2cm}
\small
\setlength{\tabcolsep}{6pt}
\renewcommand{\arraystretch}{1.1}
\resizebox{0.85\textwidth}{!}{%
\begin{tabular}{llccccccccc}
\toprule
\multirow{3}{*}{\textbf{Method}}
& \multirow{3}{*}{\textbf{Venue}}
& \multicolumn{5}{c}{\textbf{Linear evaluation}}
& \multicolumn{4}{c}{\textbf{Semi-supervised (NTU-60)}} \\
\cmidrule(lr){3-7}\cmidrule(lr){8-11}
& & \multicolumn{2}{c}{NTU-60}
& \multicolumn{2}{c}{NTU-120}
& PKU-MMD
& \multicolumn{2}{c}{X-Sub}
& \multicolumn{2}{c}{X-View} \\
\cmidrule(lr){3-4}\cmidrule(lr){5-6}
\cmidrule(lr){7-7}\cmidrule(lr){8-9}\cmidrule(lr){10-11}
& & X-Sub$\uparrow$ & X-View$\uparrow$
& X-Sub$\uparrow$ & X-Set$\uparrow$
& X-Sub$\uparrow$
& 1\%$\uparrow$ & 10\%$\uparrow$
& 1\%$\uparrow$ & 10\%$\uparrow$ \\
\midrule
\rowcolor{gray!10}
\multicolumn{11}{l}{\textit{Sensor-specific representation: one model per dataset}}\\

MAMP~\cite{mao2023mamp} & ICCV'23
& 84.9 & 89.1 & 78.6 & 79.1 & 53.8
& 66.0 & 88.0 & 68.7 & 91.5 \\

SCD-Net~\cite{wu2024scdnet} & AAAI'24
& 86.6 & 91.7 & 76.9 & 80.1 & 54.0
& 69.1 & 82.2 & 66.8 & 85.8 \\

S-JEPA~\cite{abdelfattah2024sjepa} & ECCV'24
& 85.3 & 89.8 & 79.6 & 79.9 & 53.5
& 67.5 & 88.4 & 69.1 & 91.4 \\

MacDiff~\cite{wu2024macdiff} & ECCV'24
& 86.4 & 91.0 & 79.4 & 80.2 & --
& 65.6 & 88.2 & \underline{77.3} & 92.5 \\

USDRL~\cite{weng2025usdrl} & AAAI'25
& 85.2 & 91.7 & 76.6 & 78.1 & 54.4
& 57.3 & 80.2 & 60.7 & 84.0 \\

GFP~\cite{sun2025gfp} & ICCV'25
& 85.9 & 92.0 & 79.1 & 80.3 & 56.2
& 71.8 & 88.7 & 72.9 & 92.1 \\

STARS~\cite{mehraban2026stars} & WACV'26
& 87.1 & 90.9 & 79.9 & 80.8 & 52.7
& 69.1 & 88.0 & 72.9 & 91.8 \\

AMR~\cite{sun2026amr} & CVPR'26
& 87.4 & 92.3 & 81.1 & 81.9 & 60.3
& \underline{72.2} & 89.0 & 74.4 & \underline{92.7} \\

DiMoP~\cite{gunasekara2026dimop} & TBIOM'26
& 86.8 & 91.9 & 80.7 & 81.8 & --
& 70.1 & 88.4 & 71.2 & 92.3 \\

Path-JEPA~\cite{singh2026pathjepa} & ECCV'26
& 87.6 & 92.9 & \textbf{81.5} & 81.8 & 56.7
& \underline{72.2} & \underline{89.1} & 71.8 & 92.1 \\

SLiM~\cite{do2026slim} & NeurIPS'26
& \underline{87.9} & \underline{93.2}
& \underline{81.2} & \textbf{83.6} & 59.7
& 72.1 & 88.8 & 75.8 & 91.9 \\

\midrule
\rowcolor{gray!10}
\multicolumn{11}{l}{\textit{Sensor-unified representation: skeleton data}}\\

SOfA~\cite{do2026sofa} & arXiv'26
& 87.0 & 92.6 & 79.5 & 82.0 & \underline{67.1}
& 71.6 & -- & 74.2 & -- \\

\midrule
\rowcolor{gray!10}
\multicolumn{11}{l}{\textit{Motion-skeleton unified representation}}\\

MoSk (skeleton data only) & --
& 87.2 & 93.4 & 78.8 & 81.8 & 67.8
& 77.8 & 90.7 & 79.4 & -- \\

MoSk (motion data only) & --
& 80.0 & 86.4 & 69.4 & 72.3 & 56.0
& 59.5 & 87.0 & 61.1 & -- \\

\rowcolor{Melon!12}
\textbf{MoSk (Ours)} & --
& \textbf{88.5} & \textbf{94.0}
& 81.1 & \underline{83.3} & \textbf{69.4}
& \textbf{77.0} & \textbf{91.4}
& \textbf{78.1} & \textbf{95.4} \\

\bottomrule
\end{tabular}}
\vspace{-0.3cm}
\end{table*}

\subsection{Datasets and Evaluation Protocols}
\label{sec:exp_datasets}

\noindent\textbf{Training corpus.}
The training corpus combines (i) four motion datasets such as HumanML3D~\cite{guo2022humanml3d}, KIT-ML~\cite{plappert2016kit}, MotionMillion~\cite{fan2025motionmillion}, and RoMo~\cite{zhang2026romo} and (ii) three skeleton datasets including NTU RGB+D 60/120~\cite{shahroudy2016ntu,liu2019ntu120}, PKU-MMD~II~\cite{liu2017pku}, and ETRI-Activity3D~\cite{jang2020etri}. Tab.~\ref{tab:data} summarizes the 1.47M clips spanning three native skeleton formats. Training uses neither action labels nor captions. We exclude evaluation clips and their mirrors and duplicates. In particular, NTU-120 includes NTU-60, but NTU-60 X-View and NTU-120 X-Set split sequences differently by camera views and capture setups, respectively. Training thus excludes all test clips from NTU-60 X-Sub/X-View and NTU-120 X-Sub/X-Set to avoid cross-protocol leakage.

\noindent\textbf{Understanding protocol.}
Action recognition uses joint coordinates only and reports top-1 accuracy. We follow the official NTU protocols above. PKU-MMD~II uses its official cross-subject (X-Sub) protocol. To evaluate transfer across skeleton formats, we consider both seen and unseen formats, defined by whether the format appeared during training. Seen-format benchmarks include NTU-60, NTU-120, PKU-MMD~II, and ETRI-Activity3D. Unseen-format evaluation covers two skeleton formats: (i) Kinect~v1 format (20 joints), using NW-UCLA~\cite{wang2014nwucla}, MSR-Action3D~\cite{li2010msraction3d}, and UT-Kinect~\cite{xia2012utkinect}; and (ii) OpenNI format (15 joints), using Florence-3D~\cite{seidenari2013florence} and SBU-Interaction~\cite{yun2012sbu}. Florence-3D and SBU use ten leave-one-actor-out folds and five official folds, respectively.

We assess MoSk under three settings: linear evaluation, semi-supervised evaluation, and full fine-tuning. Linear evaluation trains only the classifier using all labels on seen- and unseen-format benchmarks. Semi-supervised evaluation jointly fine-tunes MoSk and the classifier on uniformly sampled 1\% or 10\% subsets of each NTU-60 training split, without class balancing. Full fine-tuning uses the same setup with all labels. We report semi-supervised results on NTU-60 and full fine-tuning on NTU-60, NTU-120, and NW-UCLA. Results on other datasets appear in the \textit{Appendix}.

\noindent\textbf{Generation protocol.}
Generation covers unconditional and text-to-motion generation on all four motion datasets. HumanML3D and KIT-ML use the standard evaluators~\cite{guo2022humanml3d}; MotionMillion and RoMo use the released MotionMillion evaluator~\cite{fan2025motionmillion} on subsets reserved for evaluation. Fréchet Inception Distance (FID) compares real and generated motion-feature distributions. Unconditional generation also reports foot-skating rate (Skate) and mean joint jerk (Jerk) to assess sliding and smoothness. Text-to-motion generation additionally uses top-1 text--motion retrieval accuracy (R@1) and Multimodal Distance (MM-Dist) to assess semantic correspondence. Lower is better except for R@1.

\subsection{Implementation Details}
\label{sec:exp_implementation}

\noindent\textbf{Masking and augmentations.}
Joint dropout is applied with probability $0.5$ and removes a randomly sampled fraction of joints ($0$--$40\%$) across all frames. Token masking uses either uniform random masking or SLiM's strategy, each with probability $0.5$. Including joint dropout, the overall masking ratio is $90\%$ of the original skeletal-temporal tokens. Data augmentation follows SLiM, with details in the \textit{Appendix}. Clips are resampled to $T=64$ frames with speed factors in $[1/1.25,1.25]$.

\begin{figure*}[t]
\centering
\includegraphics[width=0.95\linewidth]{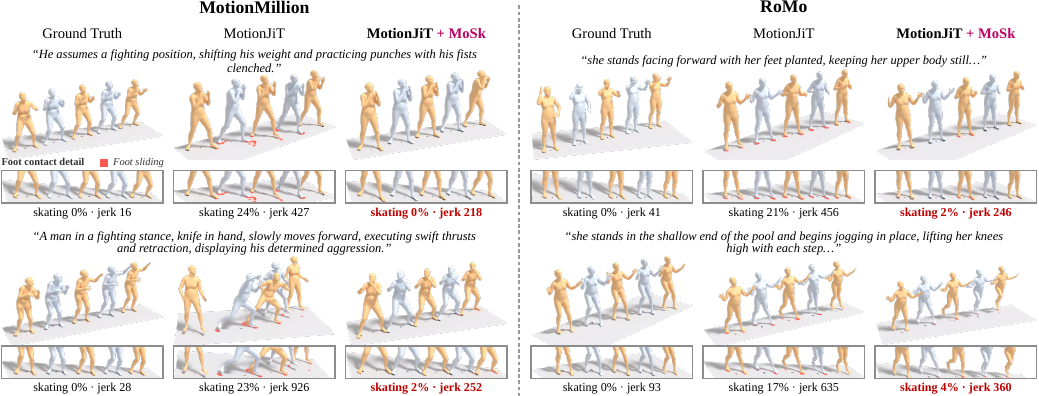}
\vspace{-0.2cm}
\caption{\textbf{Text-to-motion generation on MotionMillion and RoMo.} We compare MotionJiT with and without MoSk representation alignment using identical captions, sequence lengths, and initial noise. Insets zoom in on foot contacts, with sliding marked in red (foot height below 5\,cm and displacement above 1.67\,cm per frame). Labels report foot-skating rate and mean joint jerk (m/s$^3$).}
\label{fig:visual}
\vspace{-0.3cm}
\end{figure*}

\noindent\textbf{Architecture.}
We use a ViT encoder with 24 Transformer blocks, a hidden dimension of 1,024, and 16 attention heads. The decoder uses 4 Transformer blocks with a hidden dimension of 384 and 8 attention heads. A frozen CLIP ViT-B/32 text encoder provides joint-name embeddings.

\noindent\textbf{Training setup.}
All encoder variants are trained for 75 epochs (93,750 optimizer steps), including an initial 4-epoch linear warmup.
Each epoch consists of 1,250 optimizer steps. 
We use AdamW~\cite{loshchilov2019adamw} with a learning rate of $4\times10^{-4}$, $(\beta_1,\beta_2)=(0.9,0.95)$, and a weight decay of $0.05$.
Training uses bfloat16 (bf16) with gradient-norm clipping at $3.0$.
For generation downstream tasks, the loss weight is empirically set to $\lambda=0.5$ for all experiments.
All configurations use a total batch size of 256 across two NVIDIA B200 GPUs.
This architecture trains in about 12.8 hours and uses 23 GB of memory per GPU.

\begin{table}[tbp]
    \scriptsize
    \centering
    \caption{\textbf{Linear evaluation on unseen skeleton formats} (top-1 accuracy, \%). Joints are matched by name, with the same encoder and protocol as in Tab.~\ref{tab:linear_semi}. $^\dagger$: dataset included in SOfA's pretraining data. SkeletonMAE/MAMP results are from SOfA~\cite{do2026sofa}.}
    \vspace{-0.2cm}
    \label{tab:unseen}
    \setlength{\tabcolsep}{5pt}
    \renewcommand{\arraystretch}{1.1}
    \resizebox{0.95\columnwidth}{!}{%
    \begin{tabular}{lcccc}
        \toprule
        \multirow{2}{*}{\textbf{Method}}
        & \multicolumn{2}{c}{\textbf{Kinect v1 (20 joints)}}
        & \multicolumn{2}{c}{\textbf{OpenNI (15 joints)}} \\
        \cmidrule(lr){2-3}\cmidrule(lr){4-5}
        & \textbf{MSR-3D}
        & \textbf{UT-Kinect}
        & \textbf{Florence}
        & \textbf{SBU-Inter} \\
        \midrule
        \rowcolor{gray!10}
        \multicolumn{5}{l}{\textit{Fully supervised, target skeleton seen (reference, not ranked)}} \\
        Best reported
        & 84.5~\cite{tran2013sparse}
        & 96.5~\cite{wang2016snippets}
        & 94.4~\cite{luvizon2017harskel}
        & 98.5~\cite{wen2023istanet} \\
        \midrule
        \rowcolor{gray!10}
        \multicolumn{5}{l}{\textit{Sensor-specific representation}} \\
        SkeletonMAE~\cite{wu2023skeletonmae}
        & -- & -- & 66.8 & 73.1 \\
        MAMP~\cite{mao2023mamp}
        & -- & -- & 73.7 & 80.2 \\
        \midrule
        \rowcolor{gray!10}
        \multicolumn{5}{l}{\textit{Sensor-unified representation (skeleton data)}} \\
        SOfA~\cite{do2026sofa}
        & \textbf{82.2$^\dagger$}
        & \underline{97.0$^\dagger$}
        & \underline{91.7}
        & \textbf{97.0} \\
        \midrule
        \rowcolor{gray!10}
        \multicolumn{5}{l}{\textit{Motion-skeleton unified representation}} \\
        \rowcolor{Melon!12}
        \textbf{MoSk (Ours)}
        & \underline{81.0}
        & \textbf{98.0}
        & \textbf{96.5}
        & \underline{94.8} \\
        \bottomrule
    \end{tabular}}
    \vspace{-0.3cm}
\end{table}

\begin{table}[tbp]
    \scriptsize
    \centering
    \caption{\textbf{Full fine-tuning} (top-1 accuracy, \%; joint stream). MoSk is fine-tuned together with a linear classifier on all training clips for 100 epochs, following MAMP~\cite{mao2023mamp}. The fully supervised methods in the first block report ensemble results combining predictions from multiple skeleton streams.}
    \vspace{-0.2cm}
    \label{tab:finetune}
    \setlength{\tabcolsep}{2.5pt}
    \renewcommand{\arraystretch}{1.1}
    \resizebox{\columnwidth}{!}{%
    \begin{tabular}{llccccc}
        \toprule
        \multirow{2}{*}{\textbf{Method}} & \multirow{2}{*}{\textbf{Venue}} & \multicolumn{2}{c}{\textbf{NTU-60}} & \multicolumn{2}{c}{\textbf{NTU-120}} & \textbf{NW-UCLA} \\
        \cmidrule(lr){3-4}\cmidrule(lr){5-6}\cmidrule(lr){7-7}
        & & \textbf{X-Sub} & \textbf{X-View} & \textbf{X-Sub} & \textbf{X-Set} & \textbf{X-View} \\
        \midrule
        \rowcolor{gray!10}
        \multicolumn{7}{l}{\textit{Fully supervised, target skeleton seen (reference, not ranked)}} \\
        SkateFormer~\cite{do2024skateformer} & ECCV'24 & 93.5 & 97.8 & 89.8 & 91.4 & 98.3 \\
        ProtoGCN~\cite{liu2025protogcn} & CVPR'25 & 93.8 & 97.8 & 90.9 & 92.2 & -- \\
        Hyper-GCN~\cite{zhou2025hypergcn} & ICCV'25 & 93.7 & 97.8 & 90.9 & 92.0 & 97.6 \\
        Gamba~\cite{zhou2026gamba} & CVPR'26 & 93.4 & 97.3 & 90.1 & 91.9 & 97.3 \\
        \midrule
        \rowcolor{gray!10}
        \multicolumn{7}{l}{\textit{Sensor-specific representation}} \\
        SkeletonMAE~\cite{wu2023skeletonmae} & ICMEW'23 & 86.6 & 92.9 & 76.8 & 79.1 & -- \\
        MotionBERT~\cite{zhu2023motionbert} & ICCV'23 & 93.0 & 97.2 & -- & -- & -- \\
        MAMP~\cite{mao2023mamp} & ICCV'23 & \underline{93.1} & \underline{97.5} & \underline{90.0} & 91.3 & -- \\
        MacDiff~\cite{wu2024macdiff} & ECCV'24 & 92.7 & 97.3 & -- & -- & -- \\
        STARS~\cite{mehraban2026stars} & WACV'26 & 93.0 &\underline{97.5} & 89.9 & \underline{91.4} & -- \\
        \midrule
        \rowcolor{gray!10}
        \multicolumn{7}{l}{\textit{Motion-skeleton unified representation}} \\
        MoSk (no pre-training) & -- & 84.5 & 94.5 & 78.1 & 81.6 & 84.9 \\
        \rowcolor{Melon!12}
        \textbf{MoSk (Ours)} & -- & \textbf{94.1} & \textbf{98.7} & \textbf{90.3} & \textbf{92.4}& \textbf{97.2} \\
        \bottomrule
    \end{tabular}}
    \vspace{-0.3cm}
\end{table}

\subsection{Action Recognition}
\label{sec:exp_understanding}
We compare MoSk with dataset-specialist methods and SOfA~\cite{do2026sofa}, a model that learns across skeleton datasets.

\begin{table*}[t]
\centering
\caption{\textbf{Motion generation with MoSk representation alignment.} Each backbone is trained with and without REPA-style alignment to frozen MoSk under the same schedule. MotionJiT adapts JiT~\cite{li2026jit} with four-frame patch tokens; MotionPiT adapts PixelDiT~\cite{yu2026pixeldit} using per-frame tokens. $\mathrm{FID}^{\varnothing}$ denotes unconditional FID. Skate is reported as a fraction, and Jerk in m/s$^3$.}
\label{tab:generation}
\vspace{-0.2cm}
\small
\setlength{\tabcolsep}{5pt}
\renewcommand{\arraystretch}{1.1}
\resizebox{0.9\linewidth}{!}{%
\begin{tabular}{ll*{12}{c}}
\toprule
\multirow{3}{*}{\textbf{Dataset}}
& \multirow{3}{*}{\textbf{Alignment}}
& \multicolumn{6}{c}{\textbf{Unconditional generation}}
& \multicolumn{6}{c}{\textbf{Text-to-motion generation}} \\
\cmidrule(lr){3-8}\cmidrule(lr){9-14}
& & \multicolumn{3}{c}{MotionJiT~\cite{li2026jit}}
  & \multicolumn{3}{c}{MotionPiT~\cite{yu2026pixeldit}}
  & \multicolumn{3}{c}{MotionJiT~\cite{li2026jit}}
  & \multicolumn{3}{c}{MotionPiT~\cite{yu2026pixeldit}} \\
\cmidrule(lr){3-5}\cmidrule(lr){6-8}
\cmidrule(lr){9-11}\cmidrule(lr){12-14}
& & FID$^{\varnothing}$$\downarrow$ & Skate$\downarrow$ & Jerk$\downarrow$
  & FID$^{\varnothing}$$\downarrow$ & Skate$\downarrow$ & Jerk$\downarrow$
  & FID$\downarrow$ & R@1$\uparrow$ & MM-Dist$\downarrow$
  & FID$\downarrow$ & R@1$\uparrow$ & MM-Dist$\downarrow$ \\
\midrule
& None
& 0.692 & 0.168 & 139
& 0.326 & 0.114 & 119
& 0.052 & \textbf{0.575} & 2.65
& 0.050 & 0.571 & 2.68 \\
\rowcolor{Melon!12}
\cellcolor{white}\multirow{-2}{*}{HumanML3D~\cite{guo2022humanml3d}}
& + \textbf{MoSk (Ours)}
& \textbf{0.461} & \textbf{0.133} & \textbf{97}
& \textbf{0.310} & \textbf{0.110} & \textbf{96}
& \textbf{0.046} & \textbf{0.575} & \textbf{2.64}
& \textbf{0.049} & \textbf{0.573} & \textbf{2.66} \\
\midrule
& None
& 1.667 & 0.276 & 75
& 1.512 & \textbf{0.261} & 80
& \textbf{0.215} & 0.446 & 2.74
& 0.147 & 0.461 & \textbf{2.56} \\
\rowcolor{Melon!12}
\cellcolor{white}\multirow{-2}{*}{KIT-ML~\cite{plappert2016kit}}
& + \textbf{MoSk (Ours)}
& \textbf{1.322} & \textbf{0.272} & \textbf{70}
& \textbf{1.297} & 0.265 & \textbf{68}
& 0.218 & \textbf{0.454} & \textbf{2.70}
& \textbf{0.139} & \textbf{0.464} & 2.63 \\
\midrule
& None
& 71.71 & 0.114 & 314
& 64.31 & \textbf{0.085} & 318
& 32.19 & 0.800 & 23.87
& 30.79 & 0.809 & 23.69 \\
\rowcolor{Melon!12}
\cellcolor{white}\multirow{-2}{*}{MotionMillion~\cite{fan2025motionmillion}}
& + \textbf{MoSk (Ours)}
& \textbf{61.96} & \textbf{0.095} & \textbf{248}
& \textbf{62.95} & \textbf{0.085} & \textbf{261}
& \textbf{28.36} & \textbf{0.813} & \textbf{23.53}
& \textbf{26.97} & \textbf{0.814} & \textbf{23.54} \\
\midrule
& None
& 15.35 & 0.139 & 472
& 7.79 & 0.083 & 496
& 17.35 & 0.431 & 33.03
& 17.33 & 0.433 & 33.04 \\
\rowcolor{Melon!12}
\cellcolor{white}\multirow{-2}{*}{RoMo~\cite{zhang2026romo}}
& + \textbf{MoSk (Ours)}
& \textbf{9.68} & \textbf{0.092} & \textbf{371}
& \textbf{6.44} & \textbf{0.073} & \textbf{423}
& \textbf{15.27} & \textbf{0.435} & \textbf{32.98}
& \textbf{15.31} & \textbf{0.436} & \textbf{33.00} \\
\bottomrule
\end{tabular}}
\end{table*}

\noindent\textbf{Linear evaluation.}
Tab.~\ref{tab:linear_semi} reports the classification results on seen skeleton formats. MoSk outperforms dataset-specific specialists on both NTU-60 splits and PKU-MMD~II, reaching 94.0\% on NTU-60 X-View and 69.4\% on PKU-MMD~II, while being competitive on NTU-120. Using the combined corpus yields higher recognition accuracy than using either source alone across all five protocols. Tab.~\ref{tab:unseen} evaluates transfer to unseen Kinect~v1 and OpenNI formats. MoSk performs comparably to fully supervised references, exceeding their reported accuracies on UT-Kinect and Florence-3D. SOfA~\cite{do2026sofa} includes MSR-Action3D and UT-Kinect in its pretraining data, making Kinect~v1 a seen format for SOfA. Even so, MoSk achieves higher accuracy on UT-Kinect (98.0\% vs.\ 97.0\%). MoSk also improves on SOfA by 4.8\%-point on Florence-3D.

\noindent\textbf{Semi-supervised evaluation.}
Tab.~\ref{tab:linear_semi} also reports the classification results using only 1\% and 10\% of the NTU-60 training clips. MoSk outperforms the compared baselines in all four settings. Under NTU-60 X-Sub, MoSk achieves 77.0\% accuracy using 1\% of the training clips, outperforming AMR and Path-JEPA (both 72.2\%) by 4.8\%-point and demonstrating effective recognition with limited labeled data.

\noindent\textbf{Full fine-tuning.}
Tab.~\ref{tab:finetune} shows the classification results for full fine-tuning. MoSk surpasses fully supervised multi-stream specialists on both NTU-60 splits and NTU-120 X-Set. Compared to the methods that likewise learn skeleton representations, MoSk achieves the highest accuracy on all four NTU protocols. Without SSL, the same architecture achieves accuracies 4.2--12.3\%-point below those of fully fine-tuned MoSk across all five protocols, demonstrating the benefit of self-supervised training on the combined motion and skeleton corpus.

\begin{table*}[t]
\centering
\caption{\textbf{Data sampling ratio.} $p_K$ is the probability of sampling skeleton data. Skel.\ and Motion denote the exposure multipliers for the two data sources. Understanding uses linear evaluation (NTU: X-Sub). For generation, we evaluate MotionJiT on HumanML3D ($\mathrm{FID}^{\varnothing}$) and MotionMillion (FID and R@1) and report mean $\pm$ std.\ over three seeds for runs with MoSk alignment.}

\label{tab:mixture}
\vspace{-0.2cm}
\small
\setlength{\tabcolsep}{6pt}
\renewcommand{\arraystretch}{1.1}
\resizebox{0.8\linewidth}{!}{%
\begin{tabular}{crrcccccccc}
\toprule
\multirow{2}{*}{$p_K$} & \multicolumn{2}{c}{\textbf{Mixture}} & \multicolumn{5}{c}{\textbf{Understanding}} & \multicolumn{3}{c}{\textbf{Generation}} \\
\cmidrule(lr){2-3}\cmidrule(lr){4-8}\cmidrule(lr){9-11}
& Skel. & Motion & NTU-60 & NTU-120 & PKU-MMD & ETRI-Act & NW-UCLA & FID$^{\varnothing}$$\downarrow$ & FID$\downarrow$ & R@1$\uparrow$ \\
\midrule
\rowcolor{gray!10}\multicolumn{5}{l}{No REPA-alignment with MoSk} & & & &  0.692 & 32.19 & 0.800 \\
100\% & $138\times$ & $0.0\times$ & 87.2 & 78.8 & 67.8 & 88.0 & \textbf{95.9} & 0.498\sd{0.037} & 28.19\sd{1.20} & 0.811\sd{0.000} \\
\rowcolor{Melon!12}80\% & $110\times$ & $2.1\times$ & \textbf{88.5} & \textbf{81.1} & 69.4 & \textbf{89.7} & 94.2 & 0.461\sd{0.049} & 28.36\sd{0.28} & \textbf{0.813}\sd{0.000} \\
50\% & $69\times$ & $5.3\times$ & 88.3 & 81.0 & \textbf{69.5} & 89.0 & 93.8 & 0.518\sd{0.038} & 28.92\sd{0.20} & 0.810\sd{0.002} \\
20\% & $28\times$ & $8.4\times$ & 86.9 & 79.1 & 64.5 & 87.7 & 95.3 & 0.466\sd{0.043} & 28.48\sd{0.66} & 0.809\sd{0.002} \\
\rowcolor{Melon!8}0\% & $0\times$ & $10.6\times$ & 80.0 & 69.4 & 56.0 & 82.3 & 90.9 & \textbf{0.454}\sd{0.043} & \textbf{27.30}\sd{0.63} & \textbf{0.813}\sd{0.003} \\
\bottomrule
\end{tabular}}
\vspace{-0.3cm}
\end{table*}

\begin{table}[t]
\centering
\caption{\textbf{Ablations on training components} (linear evaluation accuracy, \%). Anatomy and Side denote joint-name embedding components in the encoder and decoder. Results are averaged by skeleton format: seen Kinect~v2 ($J_k=25$) and unseen Kinect~v1 ($J_k=20$) and OpenNI ($J_k=15$). Avg.\ denotes the mean over eight benchmarks, excluding ETRI-Activity3D.}
\label{tab:text_ablation}
\vspace{-0.2cm}
\small
\setlength{\tabcolsep}{6pt}
\renewcommand{\arraystretch}{1.1}
\resizebox{0.9\linewidth}{!}{%
\begin{tabular}{cccccccc}
\toprule
\multicolumn{2}{c}{\textbf{Encoder}} & \multicolumn{2}{c}{\textbf{Decoder}} & \textbf{Seen} & \multicolumn{2}{c}{\textbf{Unseen}} & \multirow{2}{*}{\textbf{Avg.}} \\
\cmidrule(lr){1-2}\cmidrule(lr){3-4}\cmidrule(lr){5-5}\cmidrule(lr){6-7}
Anatomy & Side & Anatomy & Side & $J_k$=25 & $J_k$=20 & $J_k$=15 & \\
\midrule
\rowcolor{Melon!12}$\checkmark$ & $\checkmark$ & $\checkmark$ & $\checkmark$ & 71.5 & 87.7 & 92.1 & \textbf{82.7} \\
\midrule
$\checkmark$ & $\times$ & $\checkmark$ & $\checkmark$ & 71.3 & 86.5 & 91.1 & 82.0 \\
$\times$ & $\checkmark$ & $\checkmark$ & $\checkmark$ & 71.5 & 87.1 & 91.0 & 82.2 \\
$\times$ & $\times$ & $\checkmark$ & $\checkmark$ & 70.7 & 86.4 & 89.2 & 81.2 \\
\midrule
$\checkmark$ & $\checkmark$ & $\checkmark$ & $\times$ & 71.0 & 86.6 & 91.4 & 82.0 \\
$\checkmark$ & $\checkmark$ & $\times$ & $\checkmark$ & 70.7 & 86.8 & 91.8 & 82.0 \\
$\checkmark$ & $\checkmark$ & $\times$ & $\times$ & 70.5 & 87.5 & 90.0 & 81.8 \\
\midrule
$\times$ & $\times$ & $\times$ & $\times$ & 71.2 & 87.8 & 91.1 & 82.4 \\
\midrule
\multicolumn{4}{l}{No random joint drop} & 71.3 & 86.5 & 91.0 & 81.9 \\
\multicolumn{4}{l}{Random masking only} & 71.6 & 86.5 & 91.4 & 82.1 \\
\multicolumn{4}{l}{Semantic-tube masking only} & 69.8 & 86.5 & 91.5 & 81.5 \\
\midrule
\multicolumn{4}{l}{Raw sequence $\mathbf{X}$ as target} & 63.1 & 83.4 & 87.7 & 76.8 \\
\bottomrule
\end{tabular}}
\vspace{-0.3cm}
\end{table}

\subsection{Motion Generation}
\label{sec:exp_generation}

We adapt JiT~\cite{li2026jit} and PixelDiT~\cite{yu2026pixeldit} for motion generation, naming the resulting backbones MotionJiT and MotionPiT, respectively. Tab.~\ref{tab:generation} compares each backbone trained with and without REPA-style alignment~\cite{yu2025repa} to the frozen MoSk teacher under matched training schedules. MoSk provides feature-level guidance only during training.

For unconditional generation, alignment improves distributional fidelity and temporal smoothness, reducing FID and jerk in all eight dataset--backbone pairs. The largest FID reduction is $9.75$ on MotionMillion. Foot skating generally decreases, with two exceptions for MotionPiT: a slight increase on KIT-ML and no change on MotionMillion.

For text-to-motion generation, the clearest gains are on MotionMillion and RoMo. Alignment improves R@1 and MM-Dist for both backbones on these datasets, with a maximum FID reduction of $3.83$ on MotionMillion. The gains are smaller on HumanML3D. On KIT-ML, some metrics improve while others worsen. Fig.~\ref{fig:visual} illustrates less foot sliding and lower jerk in paired samples generated with identical captions, lengths, and initial noise.

\subsection{Ablation Study}
\label{sec:exp_ablation}

We vary the sampling ratio between skeleton and motion data, and we ablate the masked-modeling components. Generation ablations use MotionJiT. Component ablations use MoSk-S, a smaller encoder variant described alongside an encoder-scale ablation in the \textit{Appendix}.

\noindent\textbf{Data sampling ratio.}
Tab.~\ref{tab:mixture} varies the probability $p_K$ of sampling skeleton data. Compared with skeleton-only training, adding motion data at $p_K=80\%$ improves recognition on NTU-60, NTU-120, PKU-MMD~II, and ETRI-Activity3D. Motion-only training yields the lowest mean FID in both generation settings. We use $p_K=80\%$ to balance recognition and generation performance.

\noindent\textbf{Training components.}
Tab.~\ref{tab:text_ablation} shows that using anatomy and side embeddings in both the encoder and decoder gives the highest average accuracy, with a modest gain over omitting text embeddings. 
Random joint dropout improves transfer to unseen formats, and sampling between random and semantic tube masking improves average accuracy over either strategy alone. Predicting frame-to-frame displacement instead of raw coordinates has the largest effect, improving average accuracy by 5.9\%-point.
\section{Conclusion}
\label{sec:conclusion}

We introduced MoSk, the \textit{first motion-skeleton foundation model} that learns shared representations from motion and skeleton data through self-supervised training. MoSk outperforms the compared sensor-specific methods on all four NTU fine-tuning protocols and transfers effectively to unseen skeleton formats. MoSk also improves both unconditional and text-to-motion generation, reducing FID in 15 of 16 settings across four datasets and two backbones.

\clearpage
\clearpage
\appendix

\noindent We evaluate MoSk's transfer from 3D training to video-estimated 2D skeletons and study how encoder scale and the number of sampled frames affect action recognition and motion generation. Additional quantitative and visual comparisons examine generation with and without MoSk representation alignment. Tab.~\ref{tab:supple_overview} summarizes these experiments, the construction of the NTU training corpus, the unified input format, implementation details, and limitations.

\begin{table}[h]
\scriptsize
\centering
\caption{\textbf{Overview of the \textit{Appendix}.}}
\vspace{-0.2cm}
\label{tab:supple_overview}
\setlength{\tabcolsep}{5pt}
\renewcommand{\arraystretch}{1.1}
\resizebox{1.0\linewidth}{!}{%
\begin{tabular}{cl}
\toprule
\rowcolor{Melon!12}
\textbf{Section} & \textbf{Contents} \\
\midrule
Sec.~\ref{supp:unseen} & Transfer to video-estimated 2D skeletons \\
Sec.~\ref{supp:ablation} & Encoder scale and number of sampled frames \\
Sec.~\ref{supp:dataset} & NTU training corpus and data leakage prevention \\
Sec.~\ref{supp:result} & Quantitative and visual results of motion generation \\
Sec.~\ref{supp:sup_format} & Motion and skeleton data conversion \\
Sec.~\ref{supp:implementation} & Masking and downstream protocols \\
Sec.~\ref{supp:limitations} & Limitations of corpus analysis and encoder scale \\
\bottomrule
\end{tabular}%
}
\end{table}


\section{Transfer to 2D Skeletons}
\label{supp:unseen}

\noindent\textbf{Evaluation setting.}
MoSk's self-supervised training uses only 3D skeleton and motion sequences. We evaluate MoSk for 2D skeleton-based action recognition on NTU-60 and NTU-120, using 17 COCO keypoints estimated from RGB frames by HRNet~\cite{sun2019hrnet}. This 2D 17-joint format is unseen during MoSk's 3D training. We apply the same 2D coordinate transformation to all frames within a clip and represent each joint by its coordinates $(x,y,0)$ and its anatomical name. The encoder and a linear classifier are fine-tuned together on the labeled training clips of each split. We report top-1 accuracy (\%).

\noindent\textbf{Results.}
Tab.~\ref{tab:pose2d} shows that MoSk outperforms the representation learning baselines in the second block across all four evaluation protocols, with the largest gain of 2.1\%-point on NTU-120 X-Set. It also exceeds the fully supervised references on three of the four protocols. The combined corpus of motion and skeleton data consistently outperforms either source alone, demonstrating MoSk's ability to transfer representations learned from 3D data effectively to an unseen 2D 17-joint skeleton format.

\begin{table}[t]
    \centering
    \caption{\textbf{Full fine-tuning on video-estimated 2D skeletons.} After self-supervised training on 3D data only, MoSk is fully fine-tuned on 2D skeletons with each joint represented as $(x,y,0)$. The first block lists fully supervised references. Results for BlockGCN and ProtoGCN are taken from the 10-clip evaluation reported by Peng \textit{et al.}~\cite{peng2026sbf}.}
    \label{tab:pose2d}
    \small
    \setlength{\tabcolsep}{5pt}
    \renewcommand{\arraystretch}{1.1}
    \resizebox{\linewidth}{!}{%
    \resizebox{\textwidth}{!}{%
    \begin{tabular}{llcccc}
        \toprule
        \multirow{2}{*}{\textbf{Method}} & \multirow{2}{*}{\textbf{Venue}} & \multicolumn{2}{c}{\textbf{NTU-60}} & \multicolumn{2}{c}{\textbf{NTU-120}} \\
        \cmidrule(lr){3-4}\cmidrule(lr){5-6}
        & & X-Sub & X-View & X-Sub & X-Set \\
        \midrule
        \rowcolor{gray!10}\multicolumn{6}{l}{\textit{Fully supervised, target skeleton seen (reference, not ranked)}} \\
        PoseConv3D~\cite{duan2022revisiting} & CVPR'22 & 93.7 & 96.6 & 86.0 & 89.6 \\
        BlockGCN~\cite{zhou2024blockgcn} & CVPR'24 & 93.9 & 98.2 & 87.3 & 90.7 \\
        ProtoGCN~\cite{liu2025protogcn} & CVPR'25 & 94.1 & 98.8 & 87.5 & 90.9 \\
        \midrule
        \rowcolor{gray!10}\multicolumn{6}{l}{\textit{Representation learning with 2D 17-joint inputs}} \\
        SkeletonMAE~\cite{yan2023skeletonmae} & ICCV'23 & 92.8 & 96.5 & 84.8 & 85.7 \\
        MotionBERT~\cite{zhu2023motionbert} & ICCV'23 & 93.0 & 97.2 & -- & -- \\
        MaskCLR~\cite{abdelfattah2024maskclr} & CVPR'24 & 93.9 & 97.3 & 87.4 & 89.5 \\
        \midrule
        \rowcolor{gray!10}\multicolumn{6}{l}{\textit{Representation learning with 3D motion and skeleton inputs}} \\
        MoSk (skeleton data only) & -- & 93.1 & 97.6 & 86.7 & 90.5 \\
        MoSk (motion data only) & -- & 94.1 & 97.9 & 87.1 & 90.9 \\
        \rowcolor{Melon!12}\textbf{MoSk (Ours)} & -- & \textbf{94.4} & \textbf{98.4} & \textbf{87.9} & \textbf{91.6} \\
        \bottomrule
    \end{tabular}}
    }
\end{table}


\section{Additional Ablation Study}
\label{supp:ablation}

\noindent\textbf{Encoder scale.}
Tab.~\ref{tab:scaling} shows that MoSk-L achieves the highest accuracy on NTU-60 and NTU-120, reaching 88.5\% and 81.1\%, respectively, while MoSk-XL leads on PKU-MMD~II, ETRI-Activity3D, and the unseen NW-UCLA. MoSk-S and MoSk-M achieve the lowest mean FID for unconditional and text-to-motion generation, respectively. MoSk-M, MoSk-L, and MoSk-XL tie for the highest mean R@1 (0.813).

\noindent\textbf{Number of sampled frames.}
Tab.~\ref{tab:trainig_sample} compares the number of frames sampled per input clip during MoSk training and inference. MoSk uses 64 sampled frames per input clip by default. Generation experiments use MotionJiT. For the encoder trained with the default 64-frame input, doubling the number of sampled frames at inference improves linear evaluation accuracy by 0.7\%-point on PKU-MMD and 0.9\%-point on ETRI-Act, with unchanged accuracy on NTU-60 and NW-UCLA. When training and inference clip lengths match, the highest accuracy is achieved with 64 frames on both NTU benchmarks, 96 frames on PKU-MMD and NW-UCLA, and 128 frames on ETRI-Act. Alignment with MoSk encoders trained on 64-, 96-, and 128-frame inputs improves both unconditional and text-to-motion FID relative to no alignment.

\begin{table*}[t]
\centering
\caption{\textbf{Encoder scale.} We compare MoSk variants with different encoder depths and widths on action recognition and motion generation. MoSk-L is our default model. Tab.~\ref{tab:mixture} describes the shared evaluation protocol.}

\label{tab:scaling}
\small
\setlength{\tabcolsep}{4pt}
\renewcommand{\arraystretch}{1.1}
\resizebox{1.0\textwidth}{!}{%
\begin{tabular}{lccccccccccc}
\toprule
\multirow{2}{*}{\textbf{Variant}} & \multirow{2}{*}{\textbf{Layers} $L_e$} & \multirow{2}{*}{\textbf{Width} $D$} & \multirow{2}{*}{\textbf{Params.}} & \multicolumn{5}{c}{\textbf{Understanding}} & \multicolumn{3}{c}{\textbf{Generation}} \\
\cmidrule(lr){5-9}\cmidrule(lr){10-12}
& & & & NTU-60 & NTU-120 & PKU-MMD & ETRI-Act & NW-UCLA & FID$^{\varnothing}$$\downarrow$ & FID$\downarrow$ & R@1$\uparrow$ \\
\midrule
\rowcolor{gray!8}\multicolumn{5}{l}{No REPA-alignment with MoSk} & & & & & 0.692 & 32.19 & 0.800 \\
MoSk-S & 8 & 256 & 6.6M & 81.9 & 72.5 & 60.1 & 85.4 & 91.4 & 0.422\sd{0.066} & 28.63\sd{0.02} & 0.809\sd{0.003} \\
MoSk-M & 12 & 768 & 86M & 87.6 & 79.8 & 67.1 & 88.2 & 94.6 & 0.432\sd{0.011} & \textbf{27.61}\sd{0.22} & \textbf{0.813}\sd{0.003} \\
\rowcolor{Melon!12}MoSk-L & 24 & 1024 & 304M & \textbf{88.5} & \textbf{81.1} & 69.4 & 89.7 & 94.2 & 0.461\sd{0.049} & 28.36\sd{0.28} & \textbf{0.813}\sd{0.000} \\
MoSk-XL & 32 & 1280 & 631M & 88.2 & 80.4 & \textbf{69.6} & \textbf{90.4} & \textbf{95.5} & 0.458\sd{0.042} & 27.87\sd{1.29} & \textbf{0.813}\sd{0.001} \\
\bottomrule
\end{tabular}}
\end{table*}

\begin{table*}[t]
\small
\centering
\caption{\textbf{Number of sampled frames.} We vary the number of frames sampled per training clip, with MoSk's default 64-frame setting shown in orange. We also evaluate the encoder trained with 64-frame inputs using 96 and 128 sampled frames at inference. Tab.~\ref{tab:mixture} describes the shared evaluation protocol.}

\label{tab:trainig_sample}
\setlength{\tabcolsep}{6pt}
\renewcommand{\arraystretch}{1.1}
\resizebox{0.8\textwidth}{!}{%
\begin{tabular}{ccccccccc}
\toprule
\multicolumn{2}{c}{\textbf{Sampled frames}} & \multicolumn{5}{c}{\textbf{Understanding}} & \multicolumn{2}{c}{\textbf{Generation}} \\
\cmidrule(lr){1-2}\cmidrule(lr){3-7}\cmidrule(l){8-9}
Training & Inference & NTU-60 & NTU-120 & PKU-MMD & ETRI-Act & NW-UCLA & FID$^{\varnothing}$$\downarrow$ & FID$\downarrow$ \\
\midrule
\rowcolor{gray!8}\multicolumn{5}{l}{No REPA-alignment with MoSk} &  & & 0.692 & 32.19 \\
\rowcolor{Melon!12}64 & 64 & 88.5 & 81.1 & 69.4 & 89.7 & 94.2 & 0.461\sd{0.049} & 28.36\sd{0.28} \\
64 & 96   & 88.6 & -- & 69.8 & 90.3 & 94.4 & -- & -- \\
64 & 128  & 88.5 & -- & 70.1 & 90.6 & 94.2 & -- & -- \\
96 & 96 & 87.9 & 80.4 & 70.9 & 90.7 & 95.0 & 0.522\sd{0.176} & 27.09\sd{1.05} \\
128 & 128 & 87.6 & 79.4 & 70.2 & 91.6 & 93.8 & 0.420\sd{0.033} & 27.20\sd{0.89} \\
\bottomrule
\end{tabular}}
\end{table*}


\begin{figure*}[t]
\centering
\includegraphics[width=0.6\linewidth]{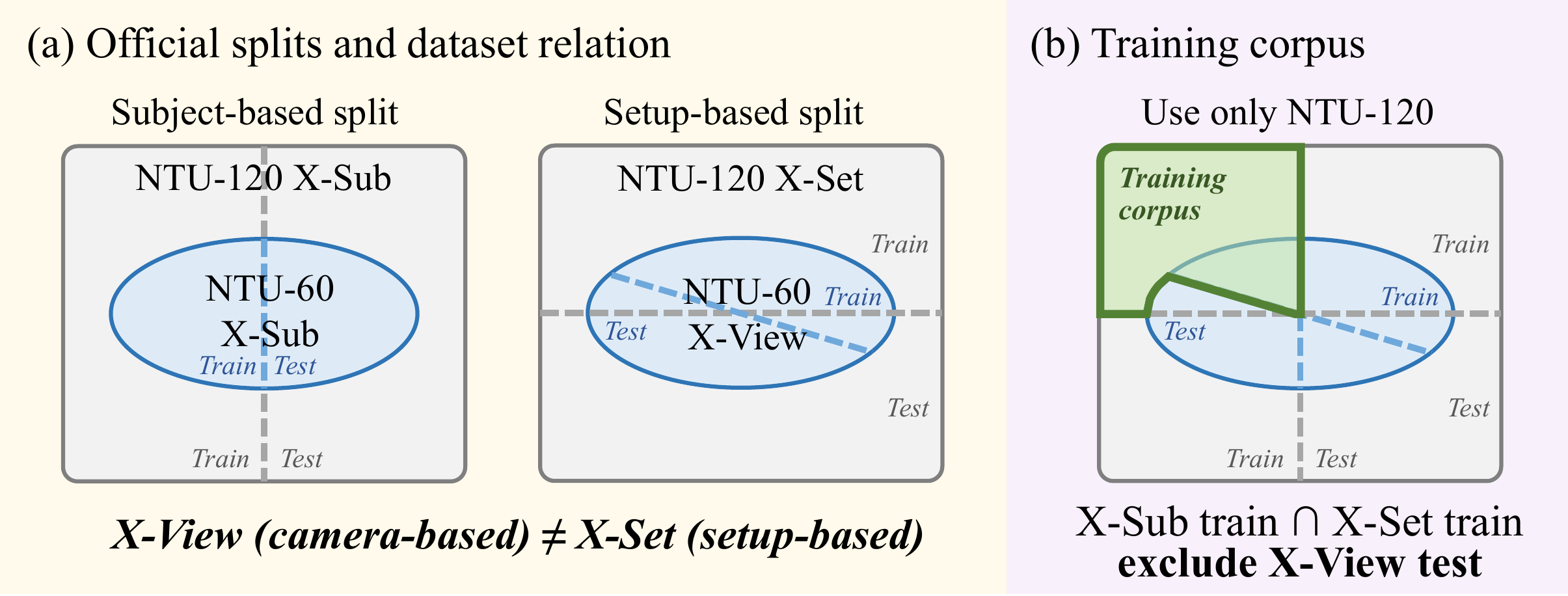}
\caption{\textbf{Construction of the NTU training corpus.} (a) NTU-60 is a subset of NTU-120, and their X-Sub protocols assign common subjects to the same training and test splits. NTU-60 X-View separates training and test clips by camera, whereas NTU-120 X-Set separates them by recording setup. (b) MoSk therefore uses only NTU-120. To prevent data leakage, we retain clips in the intersection of the NTU-120 X-Sub and X-Set training splits and exclude NTU-60 X-View test clips, yielding 18,757 training clips. The visualization is adapted from SOfA~\cite{do2026sofa}.}
\label{fig:supp_data_leakage}
\end{figure*}

\section{NTU Dataset Splits}
\label{supp:dataset}
Fig.~\ref{fig:supp_data_leakage} illustrates how we construct the NTU portion of MoSk's training corpus. Since NTU-60 is a subset of NTU-120, we use only NTU-120 to avoid duplicate clips. We retain clips in the intersection of the NTU-120 X-Sub and X-Set training splits and exclude NTU-60 X-View test clips. This yields 18,757 training clips, with verified absence of overlap with the official test sets of all four NTU evaluation protocols. SOfA~\cite{do2026sofa} handles the NTU-60 X-View overlap by removing clips from the test set to form X-View$^{*}$. MoSk instead removes overlapping clips from training, preserving the full official X-View test set.


\section{Additional Results on Generation}
\label{supp:result}

\subsection{Quantitative Results}

Tab.~\ref{tab:generation_mosk_fd} supplements Tab.~\ref{tab:generation} with additional evaluation metrics for MotionJiT and MotionPiT trained with and without REPA alignment to frozen MoSk. We report $\mathrm{FD}_{\mathrm{MoSk}}$, the Fr\'echet distance between real and generated motion distributions in the feature space of MoSk-M. This encoder differs from the one used for REPA alignment, keeping the evaluation features separate from the alignment target to support a fairer evaluation.

\noindent\textbf{Unconditional generation.}
REPA alignment improves $\mathrm{FD}_{\mathrm{MoSk}}$ and mean joint jerk across all eight dataset--backbone settings. The largest $\mathrm{FD}_{\mathrm{MoSk}}$ reduction is 14.61 for MotionJiT on HumanML3D. Foot skating decreases in six settings, with no change for MotionPiT on MotionMillion and a slight increase on KIT-ML.

\noindent\textbf{Text-to-motion generation.}
REPA alignment improves all three metrics for both backbones on HumanML3D, MotionMillion, and RoMo. On KIT-ML, foot skating decreases for both backbones, but MotionPiT's $\mathrm{FD}_{\mathrm{MoSk}}$ increases by 0.81 and MotionJiT's mean joint jerk remains unchanged.

\subsection{Visual Comparison}

All comparisons use MotionJiT with and without REPA alignment to frozen MoSk. Each generated pair shares the same sequence length and initial noise; text-to-motion pairs also share the same caption.

\begin{figure*}[t]
\centering
\includegraphics[width=1.0\linewidth]{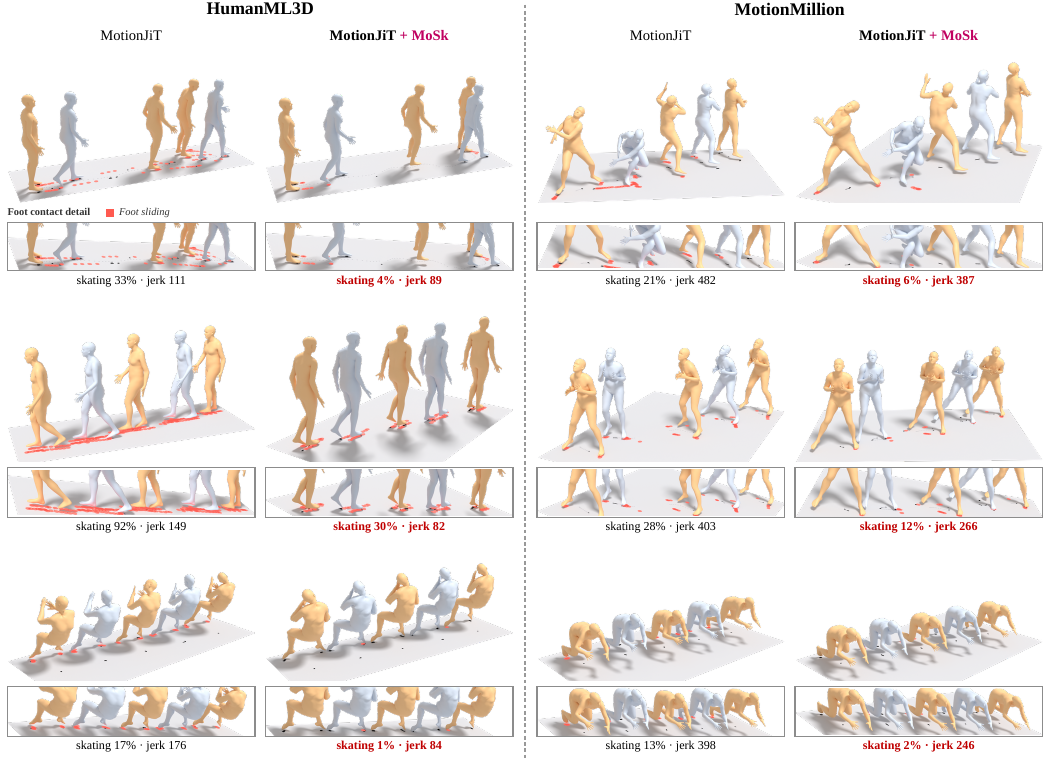}
\vspace{-0.5cm}
\caption{\textbf{Unconditional generation on HumanML3D and MotionMillion.} MotionJiT samples generated with and without MoSk representation alignment share the same sequence length and initial noise. Insets highlight foot contacts, with sliding marked in red (foot height below 5\,cm and displacement above 1.67\,cm per frame). Labels report foot-skating rate (\%) and mean joint jerk (m/s$^3$).}
\label{fig:supp_uncond1}
\end{figure*}

\begin{figure*}[t]
\centering
\includegraphics[width=1.0\linewidth]{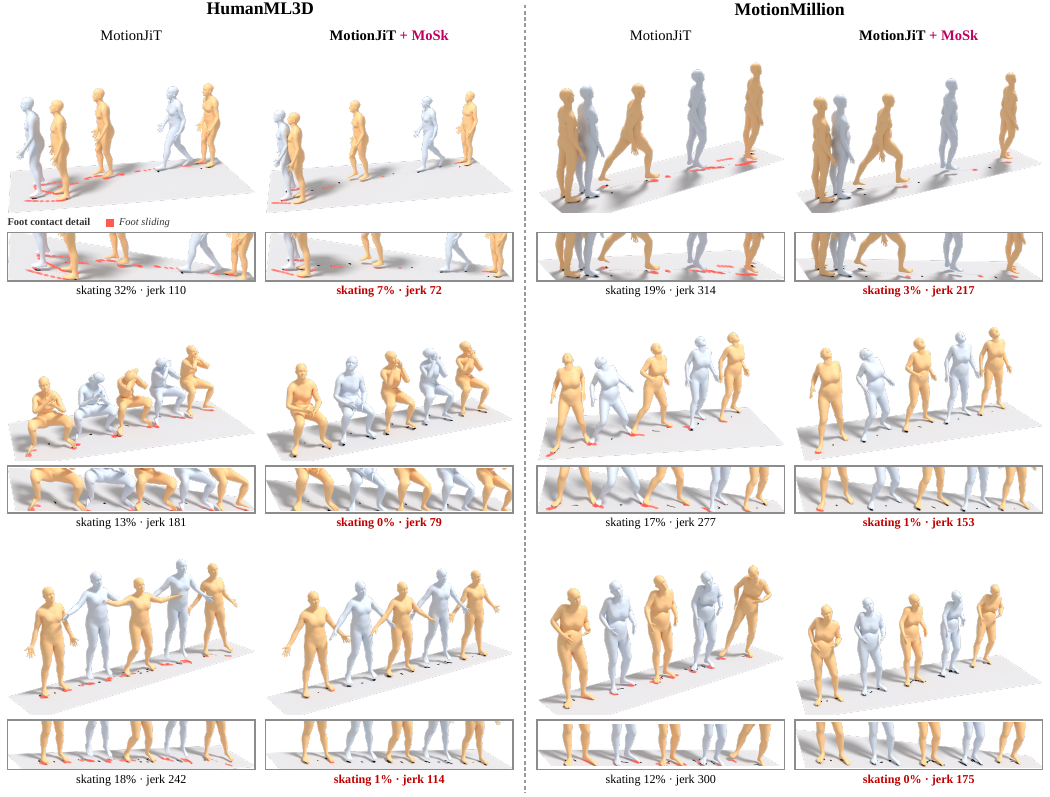}
\vspace{-0.5cm}
\caption{\textbf{Additional unconditional generation on HumanML3D and MotionMillion.} Paired samples show reduced foot sliding with MoSk. Foot-contact insets and metric labels follow the conventions in Fig.~\ref{fig:supp_uncond1}.}
\label{fig:supp_uncond2}
\end{figure*}

\noindent\textbf{Unconditional generation.}
Figs.~\ref{fig:supp_uncond1} and~\ref{fig:supp_uncond2} show samples on HumanML3D and MotionMillion. MoSk reduces foot skating and mean joint jerk in all 12 pairs shown. In the middle HumanML3D example in Fig.~\ref{fig:supp_uncond1}, the extensive red foot-sliding traces are reduced, and the foot-skating rate falls from 92\% to 30\%.

\begin{figure*}[t]
\centering
\includegraphics[width=1.0\linewidth]{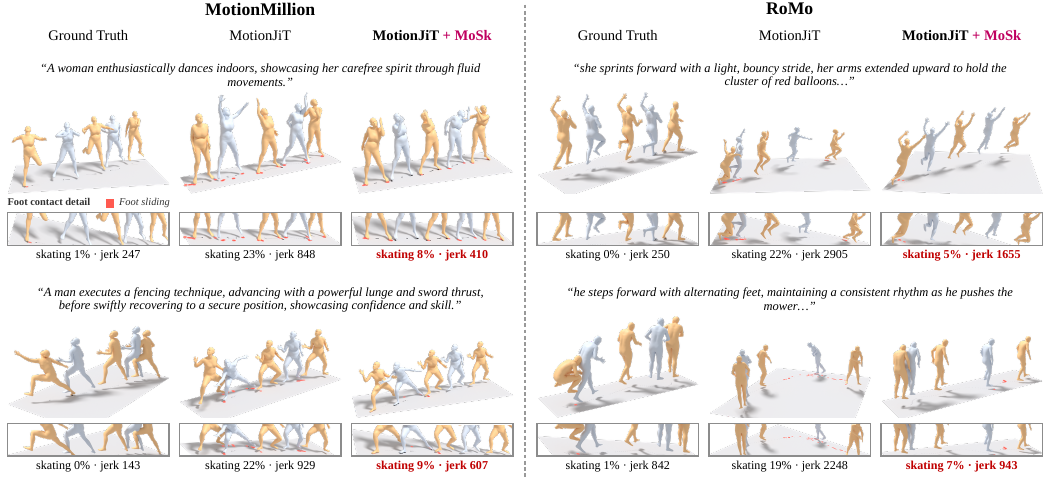}
\vspace{-0.5cm}
\caption{\textbf{Additional text-to-motion generation on MotionMillion and RoMo.} Each example shows the conditioning caption, its ground-truth reference motion, and MotionJiT generations with and without MoSk representation alignment. Insets and metric labels follow Fig.~\ref{fig:supp_uncond1}.}
\label{fig:supp_t2m}
\end{figure*}

\noindent\textbf{Text-to-motion generation.}
Fig.~\ref{fig:supp_t2m} supplements the main-paper comparison in Fig.~\ref{fig:visual} with four examples on MotionMillion and RoMo. All four pairs show less foot skating and lower mean joint jerk with MoSk. In the RoMo sprinting example, the sample generated with MoSk shows raised arms consistent with the caption, while the foot-skating rate decreases from 22\% to 5\%.

\begin{table*}[t]
\centering
\caption{\textbf{Motion generation with MoSk representation alignment.} We train each backbone with and without alignment to frozen MoSk under the same schedule. $\mathrm{FD}_{\mathrm{MoSk}}$ denotes Fr\'echet distance in the feature space of MoSk-M. Skate is the foot-skating rate (fraction), and Jerk is mean joint jerk (m/s$^3$). Unconditional Skate and Jerk are the means reported in Tab.~\ref{tab:generation}. The other metrics are averaged over a different set of training seeds. Real motion rows provide reference values for each task. Bold marks the lower mean in each pair, including ties at the displayed precision.}
\label{tab:generation_mosk_fd}
\vspace{-0.2cm}
\small
\setlength{\tabcolsep}{6pt}
\renewcommand{\arraystretch}{1.1}
\resizebox{1.0\linewidth}{!}{%
\begin{tabular}{ll*{12}{c}}
\toprule
\multirow{3}{*}{\textbf{Dataset}} & \multirow{3}{*}{\textbf{Alignment}}
& \multicolumn{6}{c}{\textbf{Unconditional generation}}
& \multicolumn{6}{c}{\textbf{Text-to-motion generation}} \\
\cmidrule(lr){3-8}\cmidrule(lr){9-14}
& & \multicolumn{3}{c}{MotionJiT~\cite{li2026jit}} & \multicolumn{3}{c}{MotionPiT~\cite{yu2026pixeldit}}
  & \multicolumn{3}{c}{MotionJiT~\cite{li2026jit}} & \multicolumn{3}{c}{MotionPiT~\cite{yu2026pixeldit}} \\
\cmidrule(lr){3-5}\cmidrule(lr){6-8}\cmidrule(lr){9-11}\cmidrule(lr){12-14}
& & $\mathrm{FD}_{\mathrm{MoSk}}\!\downarrow$ & Skate$\downarrow$ & Jerk$\downarrow$
  & $\mathrm{FD}_{\mathrm{MoSk}}\!\downarrow$ & Skate$\downarrow$ & Jerk$\downarrow$
  & $\mathrm{FD}_{\mathrm{MoSk}}\!\downarrow$ & Skate$\downarrow$ & Jerk$\downarrow$
  & $\mathrm{FD}_{\mathrm{MoSk}}\!\downarrow$ & Skate$\downarrow$ & Jerk$\downarrow$ \\
\midrule
\rowcolor{gray!10}\cellcolor{white}& Real motion
  & -- & 0.105 & 43
  & -- & 0.105 & 43
  & -- & 0.105 & 43
  & -- & 0.105 & 43 \\
& None
  & 28.15 & 0.168 & 139
  & 24.57 & 0.114 & 119
  & 22.02 & 0.189 & 198
  & 27.91 & 0.180 & 153 \\

\rowcolor{Melon!12}
\cellcolor{white}\multirow{-3}{*}{HumanML3D~\cite{guo2022humanml3d}} & + \textbf{MoSk (Ours)}
  & \textbf{13.54} & \textbf{0.133} & \textbf{97}
  & \textbf{17.71} & \textbf{0.110} & \textbf{96}
  & \textbf{19.38} & \textbf{0.182} & \textbf{178}
  & \textbf{26.23} & \textbf{0.164} & \textbf{151} \\
\midrule
\rowcolor{gray!10}\cellcolor{white}& Real motion
  & -- & 0.244 & 58
  & -- & 0.244 & 58
  & -- & 0.244 & 58
  & -- & 0.244 & 58 \\
& None
  & 1.45 & 0.276 & 75
  & 1.40 & \textbf{0.261} & 80
  & 2.98 & 0.332 & \textbf{139}
  & \textbf{2.78} & 0.309 & 119 \\
\rowcolor{Melon!12}
\cellcolor{white}\multirow{-3}{*}{KIT-ML~\cite{plappert2016kit}} & + \textbf{MoSk (Ours)}
  & \textbf{1.10} & \textbf{0.272} & \textbf{70}
  & \textbf{1.23} & 0.265 & \textbf{68}
  & \textbf{2.63} & \textbf{0.313} & \textbf{139}
  & 3.59 & \textbf{0.289} & \textbf{114} \\
\midrule
\rowcolor{gray!10}\cellcolor{white}& Real motion
  & -- & 0.054 & 73
  & -- & 0.054 & 73
  & -- & 0.054 & 73
  & -- & 0.054 & 73 \\
& None
  & 36.8 & 0.114 & 314
  & 25.5 & \textbf{0.085} & 318
  & 37.3 & 0.111 & 499
  & 40.0 & 0.110 & 513 \\
\rowcolor{Melon!12}
\cellcolor{white}\multirow{-3}{*}{MotionMillion~\cite{fan2025motionmillion}} & + \textbf{MoSk (Ours)}
  & \textbf{23.6} & \textbf{0.095} & \textbf{248}
  & \textbf{21.0} & \textbf{0.085} & \textbf{261}
  & \textbf{32.9} & \textbf{0.095} & \textbf{441}
  & \textbf{38.3} & \textbf{0.109} & \textbf{489} \\
\midrule
\rowcolor{gray!10}\cellcolor{white}& Real motion
  & -- & 0.045 & 169
  & -- & 0.045 & 169
  & -- & 0.044 & 167
  & -- & 0.044 & 167 \\
& None
  & 34.9 & 0.139 & 472
  & 26.9 & 0.083 & 496
  & 39.9 & 0.129 & 787
  & 39.9 & 0.131 & 748 \\
\rowcolor{Melon!12}
\cellcolor{white}\multirow{-3}{*}{RoMo~\cite{zhang2026romo}} & + \textbf{MoSk (Ours)}
  & \textbf{22.4} & \textbf{0.092} & \textbf{371}
  & \textbf{21.2} & \textbf{0.073} & \textbf{423}
  & \textbf{34.8} & \textbf{0.109} & \textbf{703}
  & \textbf{38.7} & \textbf{0.122} & \textbf{731} \\
\bottomrule
\end{tabular}}
\end{table*}

\section{Unified Input Format}
\label{supp:sup_format}

\begin{figure*}[t]
\centering
\includegraphics[width=\textwidth]{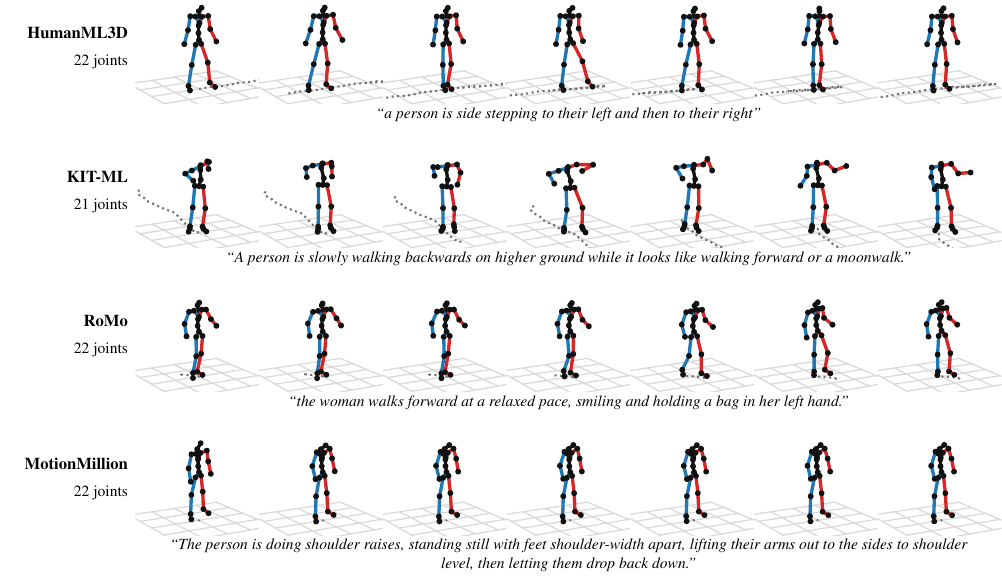}
\caption{\textbf{Inputs from motion datasets.} Each row shows seven frames after conversion from motion features to MoSk's unified input format, together with the corresponding text description. MoSk uses only joint positions during self-supervised training. Red and blue mark anatomical left and right, dotted lines trace pelvis trajectories, and grid spacing is 0.5\,m.}
\label{fig:supp_mocap}
\end{figure*}

\begin{figure*}[t]
\centering
\includegraphics[width=\textwidth]{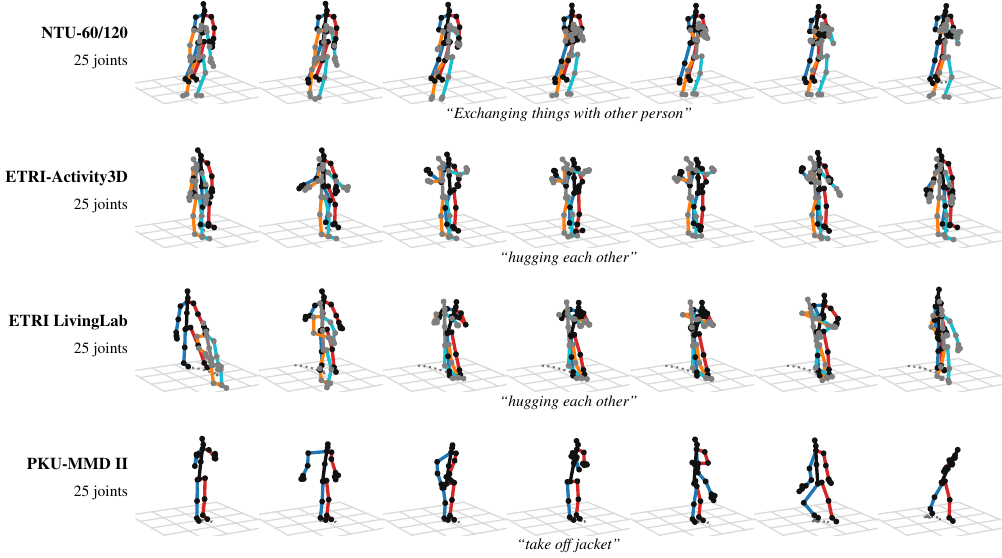}
\caption{\textbf{Kinect v2 inputs.} The examples show joint sequences after conversion to MoSk's unified input format, retaining the 25-joint layout and relative positions between people. Orange and teal mark anatomical left and right for the second person. ETRI LivingLab is shown for reference and excluded from MoSk's self-supervised training. Other visualization conventions follow Fig.~\ref{fig:supp_mocap}.}
\label{fig:supp_kinect}
\end{figure*}

\begin{figure*}[t]
\centering
\includegraphics[width=\textwidth]{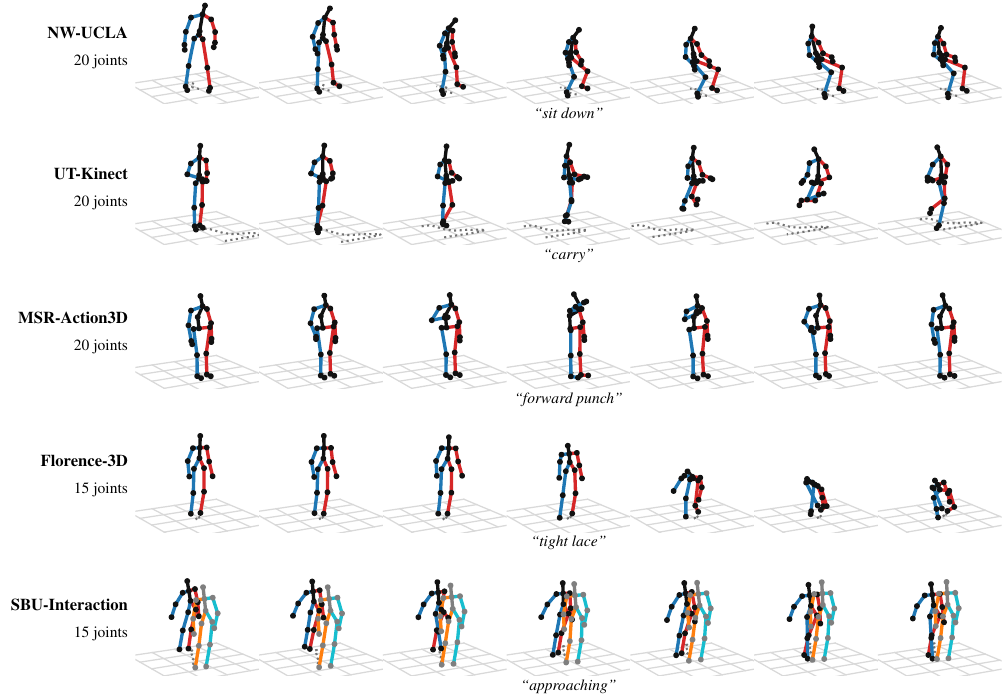}
\caption{\textbf{Inputs with unseen skeleton layouts.} The examples show joint sequences after conversion to MoSk's unified input format, retaining the original Kinect v1 (20 joints) and OpenNI (15 joints) layouts. These datasets are used only for downstream evaluation, and their joint layouts are unseen during MoSk's self-supervised training. Visualization conventions follow Figs.~\ref{fig:supp_mocap} and~\ref{fig:supp_kinect}.}
\label{fig:supp_unseen}
\end{figure*}

We store each clip offline as a $T\times M\times J\times 3$ tensor of joint positions in meters, retaining the source joint layout and body size. $T$ denotes the number of frames, $M\leq 2$ the number of people, and $J$ the number of joints. Clips are stored at their original frame rates and resampled to 20\,fps during data loading. Figs.~\ref{fig:supp_mocap}--\ref{fig:supp_unseen} illustrate the processed joint sequences across datasets.

\noindent\textbf{Feature conversion for motion datasets.}
Motion datasets including HumanML3D, KIT-ML, MotionMillion, and RoMo provide per-frame features that need conversion to 3D joint positions using their reference implementations. Each sequence contains one person ($M=1$). For HumanML3D~\cite{guo2022humanml3d} (263-D) and KIT-ML~\cite{plappert2016kit} (251-D), root heading and horizontal position are recovered by accumulating their per-frame changes, while root height is read directly from the features. Local non-root joints are rotated around the vertical axis using the recovered heading and shifted by the horizontal root translation. Prepending the root position yields $T\times22\times3$ joint positions for HumanML3D and $T\times21\times3$ for KIT-ML. Mirror correction is applied to HumanML3D motions sourced from HumanAct12~\cite{guo2020action2motion}. For KIT-ML, we correct mirroring and swap the left and right joint labels. For MotionMillion~\cite{fan2025motionmillion} and RoMo~\cite{zhang2026romo}, heading and horizontal root trajectories are recovered from their released 272-D per-frame motion features, yielding $T\times22\times3$ joint positions for each dataset.

\noindent\textbf{Coordinate frame and units.}
Both motion and skeleton data are standardized to the HumanML3D~\cite{guo2022humanml3d} convention, with $+y$ pointing up and the floor at $y=0$. Tab.~\ref{tab:sup_format} summarizes the dataset-specific conversions, including floor-plane estimation from tracked foot positions using RANSAC~\cite{fischler1981random}. The fitted plane normal defines the up direction if at least 30\% of the points are inliers and it lies within $15^\circ$ of the spine axis. Otherwise, the spine axis is used. Florence-3D coordinates are converted from millimeters to meters, and SBU-Interaction's coordinate normalization is reversed.

\noindent\textbf{Facing direction and origin.}
Both motion and skeleton data undergo the same facing-direction and origin alignment. For each sampled training clip, the first person in the stored data serves as the alignment reference. We estimate this person's initial orientation by averaging the horizontal hip and shoulder directions over approximately 0.33\,s at the start of the observed motion. A single transform aligns anatomical left with $+x$ and forward with $+z$, and places the reference person's pelvis at $x=z=0$ at the first observed frame. The same transform is applied to all people and frames, preserving their relative positions.

\begin{table*}[t]
\centering
\caption{\textbf{Dataset conversion to a common coordinate frame.} Layout specifies the source joint format and count. Dataset-specific rules align the up direction to $+y$ and the floor to $y=0$. Native fps lists the original frame rates of the stored clips, before resampling to 20\,fps during data loading.}
\label{tab:sup_format}
\vspace{-0.2cm}
\small
\setlength{\tabcolsep}{8pt}
\renewcommand{\arraystretch}{1.1}
\resizebox{0.9\linewidth}{!}{%
\begin{tabular}{lllr}
\toprule
Dataset & Layout (joints) & Up direction and floor & Native fps \\
\midrule
NTU-60/120~\cite{shahroudy2016ntu,liu2019ntu120} & Kinect v2 (25) & Floor plane per setup and camera & 30 \\
ETRI-Activity3D, LivingLab~\cite{jang2020etri} & Kinect v2 (25) & Floor plane per tracked sensor session & 30 \\
PKU-MMD II~\cite{liu2017pku} & Kinect v2 (25) & Floor plane per camera session & 30 \\
NW-UCLA~\cite{wang2014nwucla} & Kinect v1 (20) & Floor plane per view and subject & 15 \\
UT-Kinect~\cite{xia2012utkinect} & Kinect v1 (20) & One floor plane for the static camera & 30 \\
MSR-Action3D~\cite{li2010msraction3d} & Kinect v1 (20) & Ankle-to-shoulder axis, floor at lowest foot & 15 \\
Florence-3D~\cite{seidenari2013florence} & OpenNI (15) & Spine axis, floor 7\,cm below median lowest ankle & 20 \\
SBU-Interaction~\cite{yun2012sbu} & OpenNI (15) & Spine axis, floor 7\,cm below median lowest ankle & 15 \\
\midrule
HumanML3D~\cite{guo2022humanml3d} & SMPL (22) & Provided & 20 \\
KIT-ML~\cite{plappert2016kit} & MMM (21) & Provided & 7.5 / 12.5 / 15 \\
RoMo~\cite{zhang2026romo} & SMPL (22) & Provided, with 0.156\,m floor offset removed & 30 \\
MotionMillion~\cite{fan2025motionmillion} & SMPL (22) & Provided & 30 \\
\bottomrule
\end{tabular}}
\end{table*}

\section{Implementation Details}
\label{supp:implementation}

\subsection{Masking and Augmentations}

We sample two 64-frame training clips from each source sequence, with part of the original motion included in both clips. Both clips share the same left--right flip decision, while playback speeds, bone-length scaling factors, and small 3D rotations are sampled independently for each clip. Playback speeds follow a log-uniform distribution over $[1/1.25,1.25]$. Every four consecutive frames form one temporal patch per joint, giving 16 patches per joint in each clip. Joint dropout is applied with probability 0.5 per training batch, removing a random subset containing 0--40\% of the joints from every frame. Each clip then uses either \emph{random masking} or \emph{semantic tube masking} from SLiM~\cite{do2026slim}, with equal probability. Random masking selects individual tokens uniformly at random, whereas semantic tube masking selects connected joints of the trunk or a limb over consecutive temporal patches. Joint dropout and token masking together hide approximately 90\% of the original non-padding tokens, with at least one token remaining visible. The decoder reconstructs the complete original joint set, including joints removed by dropout.

\subsection{Action Understanding}

For an input sequence $S_k$ from the $k$-th dataset, we apply the preprocessing operator $\mathcal{A}_k$ to obtain $\mathbf{X}=\mathcal{A}_k(S_k)$. We average the final output tokens of the MoSk encoder $\mathcal{E}_{M}$ to obtain one feature per clip,
\[
\mathbf{h}=\operatorname{MeanPool}\!\left(\mathcal{E}_{M}(\mathbf{X})\right)\in\mathbb{R}^{1\times C_e},
\]
where $C_e$ is the encoder width. A linear classification head maps the clip feature $\mathbf{h}$ to action scores. We train the classifier with cross-entropy and report top-1 accuracy.

\noindent\textbf{Linear evaluation.}
Following MAMP~\cite{mao2023mamp}, we freeze the MoSk encoder $\mathcal{E}_{M}$ and train only the linear classification head using all labeled training clips.

\noindent\textbf{Semi-supervised learning.}
The semi-supervised results reported in Tab.~\ref{tab:linear_semi} are obtained by jointly fine-tuning the MoSk encoder and linear classification head using MAMP's end-to-end fine-tuning recipe. For each NTU-60 training split, 1\% or 10\% of the clips are sampled uniformly at random from the entire split, without class balancing.

\noindent\textbf{Full fine-tuning.}
We jointly fine-tune the MoSk encoder and linear classification head using all labeled training clips of each evaluation split. Training follows the same MAMP end-to-end fine-tuning recipe as in the semi-supervised setting.

\subsection{Motion Generation}

\noindent\textbf{Generator inputs.}
MotionJiT~\cite{li2026jit} and MotionPiT~\cite{yu2026pixeldit} use the datasets' motion features: 263-D for HumanML3D, 251-D for KIT-ML, and 272-D for MotionMillion and RoMo. Each feature channel is standardized using training-set statistics. MotionJiT groups four consecutive frames into one token, whereas MotionPiT assigns one token to each frame. Both models predict clean motion features from noisy inputs. For text-to-motion generation, frozen CLIP~\cite{radford2021clip} text encoders provide caption embeddings.

\noindent\textbf{MoSk representation alignment.}
The 3D joint positions of each clean training clip are resampled to 20\,fps. Each non-overlapping 64-frame segment undergoes the coordinate alignment described in Sec.~\ref{supp:sup_format} before being encoded by frozen MoSk. MoSk processes all valid tokens and produces one output token per joint for every four input frames. Averaging the final-layer tokens across joints gives 16 temporal target features per segment. When the generator inputs use a frame rate other than 20\,fps, we linearly interpolate neighboring MoSk features to obtain a target at the center time of each four-frame generator group.

Each MoSk target feature channel is standardized over the valid temporal patches within its training clip. Hidden states from the fourth generator block are mapped to the MoSk feature space by a trainable projector. For MotionPiT, the four frame tokens in each group are averaged before projection. The projected features are aligned with the MoSk targets through $\mathcal{L}_{\mathrm{REPA}}$, with alignment weight $\lambda=0.5$ in Eq.~\ref{eq:generative_loss}. MoSk and the projector are used only during generator training.

\noindent\textbf{Evaluation.}
HumanML3D and KIT-ML are evaluated using the dataset-specific text-motion evaluators of Guo \textit{et al.}~\cite{guo2022humanml3d}. For both MotionMillion and RoMo, we use the released MotionMillion evaluator~\cite{fan2025motionmillion}.


\section{Limitations}
\label{supp:limitations}

\noindent\textbf{Training corpus.}
Our corpus ablations compare training on motion data, skeleton data, and their combination, without isolating the contribution of individual datasets such as NTU or RoMo. The extent to which each dataset benefits action recognition and motion generation therefore remains unclear. Training on individual datasets or excluding one dataset at a time would help quantify these effects.

\noindent\textbf{Encoder scale.}
Our default encoder, MoSk-L, has 304M parameters, making fine-tuning demanding in both computation and memory. This limits its practicality under constrained resources. Although smaller variants, MoSk-S and MoSk-M, remain effective for REPA alignment, reducing encoder size can lower action recognition accuracy (Tab.~\ref{tab:scaling}). Reducing the cost of downstream adaptation while preserving performance across tasks remains a direction for future work.

\clearpage

{
    \small
    \bibliographystyle{ieeenat_fullname}
    \bibliography{main}

@inproceedings{he2022mae,
  title={Masked autoencoders are scalable vision learners},
  author={He, Kaiming and Chen, Xinlei and Xie, Saining and Li, Yanghao and Doll{\'a}r, Piotr and Girshick, Ross},
  booktitle={2022 IEEE/CVF conference on computer vision and pattern recognition (CVPR)},
  pages={15979--15988},
  year={2022},
  organization={IEEE}
}

@article{oquab2023dinov2,
  title={Dinov2: Learning robust visual features without supervision},
  author={Oquab, Maxime and Darcet, Timoth{\'e}e and Moutakanni, Th{\'e}o and Vo, Huy and Szafraniec, Marc and Khalidov, Vasil and Fernandez, Pierre and Haziza, Daniel and Massa, Francisco and El-Nouby, Alaaeldin and others},
  author={Oquab, Maxime and Darcet, Timoth{\'e}e and Moutakanni, Th{\'e}o and Vo, Huy and Szafraniec, Marc and Khalidov, Vasil and Fernandez, Pierre and Haziza, Daniel and Massa, Francisco and El-Nouby, Alaaeldin and others},
  journal={arXiv preprint arXiv:2304.07193},
  year={2023}
}

@inproceedings{yu2025repa,
  title={Representation Alignment for Generation: Training Diffusion Transformers Is Easier Than You Think},
  author={Sihyun Yu and Sangkyung Kwak and Huiwon Jang and Jongheon Jeong and Jonathan Huang and Jinwoo Shin and Saining Xie},
  year={2025},
  booktitle={International Conference on Learning Representations},
}

@inproceedings{zheng2026rae,
  title={Diffusion transformers with representation autoencoders},
  author={Zheng, Boyang and Ma, Nanye and Tong, Shengbang and Xie, Saining},
  booktitle={International Conference on Learning Representations},
  volume={2026},
  pages={35791--35820},
  year={2026}
}

@inproceedings{caron2021dino,
  title={Emerging properties in self-supervised vision transformers},
  author={Caron, Mathilde and Touvron, Hugo and Misra, Ishan and J{\'e}gou, Herv{\'e} and Mairal, Julien and Bojanowski, Piotr and Joulin, Armand},
  booktitle={2021 IEEE/CVF international conference on computer vision (ICCV)},
  pages={9630--9640},
  year={2021},
  organization={IEEE}
}

@inproceedings{guo2022humanml3d,
  title={Generating diverse and natural 3d human motions from text},
  author={Guo, Chuan and Zou, Shihao and Zuo, Xinxin and Wang, Sen and Ji, Wei and Li, Xingyu and Cheng, Li},
  booktitle={2022 IEEE/CVF Conference on Computer Vision and Pattern Recognition (CVPR)},
  pages={5142--5151},
  year={2022},
  organization={IEEE}
}

@article{plappert2016kit,
  title={The kit motion-language dataset},
  author={Plappert, Matthias and Mandery, Christian and Asfour, Tamim},
  journal={Big data},
  volume={4},
  number={4},
  pages={236--252},
  year={2016},
  publisher={SAGE Publications, Sage CA: Los Angeles, CA}
}

@inproceedings{zhang2026romo,
  title={Romo: A large-scale, richly organized dataset and semantic taxonomy for human motion generation},
  author={Zhang, Jiahao and Liu, Joseph and Lee, Young-Yoon and Moon, Seonghyeon and Zordan, Victor and Tevet, Guy and Liu, C Karen and Gould, Stephen and Jacob, Oren and Jiang, Haomiao and others},
  booktitle={Proceedings of the IEEE/CVF Conference on Computer Vision and Pattern Recognition},
  pages={16408--16419},
  year={2026}
}

@inproceedings{fan2025motionmillion,
  title={Go to zero: Towards zero-shot motion generation with million-scale data},
  author={Fan, Ke and Lu, Shunlin and Dai, Minyue and Yu, Runyi and Xiao, Lixing and Dou, Zhiyang and Dong, Junting and Ma, Lizhuang and Wang, Jingbo},
  booktitle={2025 IEEE/CVF International Conference on Computer Vision (ICCV)},
  pages={13336--13348},
  year={2025},
  organization={IEEE}
}

@inproceedings{shahroudy2016ntu,
  title={Ntu rgb+ d: A large scale dataset for 3d human activity analysis},
  author={Shahroudy, Amir and Liu, Jun and Ng, Tian-Tsong and Wang, Gang},
  booktitle={Proceedings of the IEEE conference on computer vision and pattern recognition},
  pages={1010--1019},
  year={2016}
}

@article{liu2019ntu120,
  title={Ntu rgb+ d 120: A large-scale benchmark for 3d human activity understanding},
  author={Liu, Jun and Shahroudy, Amir and Perez, Mauricio and Wang, Gang and Duan, Ling-Yu and Kot, Alex C},
  journal={IEEE transactions on pattern analysis and machine intelligence},
  volume={42},
  number={10},
  pages={2684--2701},
  year={2019},
  publisher={IEEE}
}

@article{liu2017pku,
  title={Pku-mmd: A large scale benchmark for continuous multi-modal human action understanding},
  author={Liu, Chunhui and Hu, Yueyu and Li, Yanghao and Song, Sijie and Liu, Jiaying},
  journal={arXiv preprint arXiv:1703.07475},
  year={2017}
}

@article{do2026sofa,
  title={One for All: Generalist Foundation Model for Cross-Sensor Skeleton Representation Learning},
  author={Do, Jeonghyeok and Chen, Yun and Kim, Munchurl},
  journal={arXiv preprint arXiv:2609.07078},
  year={2026}
}

@inproceedings{chen2020simclr,
  title={A simple framework for contrastive learning of visual representations},
  author={Chen, Ting and Kornblith, Simon and Norouzi, Mohammad and Hinton, Geoffrey},
  booktitle={International conference on machine learning},
  pages={1597--1607},
  year={2020},
  organization={PmLR}
}

@inproceedings{he2020momentum,
  title={Momentum contrast for unsupervised visual representation learning},
  author={He, Kaiming and Fan, Haoqi and Wu, Yuxin and Xie, Saining and Girshick, Ross},
  booktitle={2020 IEEE/CVF conference on computer vision and pattern recognition (CVPR)},
  pages={9726--9735},
  year={2020},
  organization={IEEE}
}

@article{tong2022videomae,
  title={Videomae: Masked autoencoders are data-efficient learners for self-supervised video pre-training},
  author={Tong, Zhan and Song, Yibing and Wang, Jue and Wang, Limin},
  journal={Advances in neural information processing systems},
  volume={35},
  pages={10078--10093},
  year={2022}
}

@article{simeoni2025dinov3,
  title={Dinov3},
  author={Sim{\'e}oni, Oriane and Vo, Huy V and Seitzer, Maximilian and Baldassarre, Federico and Oquab, Maxime and Jose, Cijo and Khalidov, Vasil and Szafraniec, Marc and Yi, Seungeun and Ramamonjisoa, Micha{\"e}l and others},
  journal={arXiv preprint arXiv:2508.10104},
  year={2025}
}

@inproceedings{assran2023ijepa,
  title={Self-supervised learning from images with a joint-embedding predictive architecture},
  author={Assran, Mahmoud and Duval, Quentin and Misra, Ishan and Bojanowski, Piotr and Vincent, Pascal and Rabbat, Michael and LeCun, Yann and Ballas, Nicolas},
  booktitle={2023 IEEE/CVF Conference on Computer Vision and Pattern Recognition (CVPR)},
  pages={15619--15629},
  year={2023},
  organization={IEEE}
}

@article{bardes2024vjepa,
  title={Revisiting feature prediction for learning visual representations from video},
  author={Bardes, Adrien and Garrido, Quentin and Ponce, Jean and Chen, Xinlei and Rabbat, Michael and LeCun, Yann and Assran, Mahmoud and Ballas, Nicolas},
  journal={arXiv preprint arXiv:2404.08471},
  year={2024}
}

@inproceedings{yao2025vavae,
  title={Reconstruction vs. generation: Taming optimization dilemma in latent diffusion models},
  author={Yao, Jingfeng and Yang, Bin and Wang, Xinggang},
  booktitle={2025 IEEE/CVF Conference on Computer Vision and Pattern Recognition (CVPR)},
  pages={15703--15712},
  year={2025},
  organization={IEEE}
}

@inproceedings{leng2025repae,
  title={Repa-e: Unlocking vae for end-to-end tuning with latent diffusion transformers},
  author={Leng, Xingjian and Singh, Jaskirat and Hou, Yunzhong and Xing, Zhenchang and Xie, Saining and Zheng, Liang},
  booktitle={2025 IEEE/CVF International Conference on Computer Vision (ICCV)},
  pages={18262--18272},
  year={2025},
  organization={IEEE}
}

@inproceedings{bi2026vfmvae,
  title     = {Vision Foundation Models Can Be Good Tokenizers for Latent Diffusion Models},
  author    = {Bi, Tianci and Zhang, Xiaoyi and Lu, Yan and Zheng, Nanning},
  booktitle = {Proceedings of the IEEE/CVF Conference on Computer Vision and Pattern Recognition (CVPR)},
  year      = {2026}
}

@article{li2021crossclr,
  title={3d human action representation learning via cross-view consistency pursuit},
  author={Li, Linguo and Wang, Minsi and Ni, Bingbing and Wang, Hang and Yang, Jiancheng and Zhang, Wenjun},
  journal={arXiv preprint arXiv:2104.14466},
  year={2021}
}

@inproceedings{lin2023actionlet,
  title={Actionlet-dependent contrastive learning for unsupervised skeleton-based action recognition},
  author={Lin, Lilang and Zhang, Jiahang and Liu, Jiaying},
  booktitle={2023 IEEE/CVF Conference on Computer Vision and Pattern Recognition (CVPR)},
  pages={2363--2372},
  year={2023},
  organization={IEEE}
}

@inproceedings{mao2023mamp,
  title={Masked motion predictors are strong 3d action representation learners},
  author={Mao, Yunyao and Deng, Jiajun and Zhou, Wengang and Fang, Yao and Ouyang, Wanli and Li, Houqiang},
  booktitle={2023 IEEE/CVF International Conference on Computer Vision (ICCV)},
  pages={10147--10157},
  year={2023},
  organization={IEEE}
}

@inproceedings{abdelfattah2024sjepa,
  title={S-jepa: A joint embedding predictive architecture for skeletal action recognition},
  author={Abdelfattah, Mohamed and Alahi, Alexandre},
  booktitle={European Conference on Computer Vision},
  pages={367--384},
  year={2024},
  organization={Springer}
}

@inproceedings{do2026slim,
  title={Less is More: Compact-Token Masked Feature Prediction for Skeleton Representation Learning},
  author={Do, Jeonghyeok and Chen, Yun and Youk, Geunhyuk and Kim, Munchurl},
  booktitle={Advances in Neural Information Processing Systems (NeurIPS)},
  year={2026}
}

@inproceedings{zhu2023motionbert,
  title={Motionbert: A unified perspective on learning human motion representations},
  author={Zhu, Wentao and Ma, Xiaoxuan and Liu, Zhaoyang and Liu, Libin and Wu, Wayne and Wang, Yizhou},
  booktitle={2023 IEEE/CVF International Conference on Computer Vision (ICCV)},
  pages={15039--15053},
  year={2023},
  organization={IEEE}
}

@inproceedings{weng2025usdrl,
  title={Usdrl: Unified skeleton-based dense representation learning with multi-grained feature decorrelation},
  author={Weng, Wanjiang and Wang, Hongsong and Wang, Junbo and He, Lei and Xie, Guo-Sen},
  booktitle={Proceedings of the AAAI conference on artificial intelligence},
  volume={39},
  number={8},
  pages={8332--8340},
  year={2025}
}

@inproceedings{chen2023mld,
  title={Executing your commands via motion diffusion in latent space},
  author={Chen, Xin and Jiang, Biao and Liu, Wen and Huang, Zilong and Fu, Bin and Chen, Tao and Yu, Gang},
  booktitle={2023 IEEE/CVF Conference on Computer Vision and Pattern Recognition (CVPR)},
  pages={18000--18010},
  year={2023},
  organization={IEEE}
}

@inproceedings{zhang2023t2mgpt,
  title={T2M-GPT: Generating Human Motion from Textual Descriptions with Discrete Representations},
  author={Zhang, Jianrong and Zhang, Yangsong and Cun, Xiaodong and Huang, Shaoli and Zhang, Yong and Zhao, Hongwei and Lu, Hongtao and Shen, Xi},
  booktitle={Proceedings of the IEEE/CVF Conference on Computer Vision and Pattern Recognition (CVPR)},
  year={2023},
}

@inproceedings{guo2024momask,
  title={Momask: Generative masked modeling of 3d human motions},
  author={Guo, Chuan and Mu, Yuxuan and Javed, Muhammad Gohar and Wang, Sen and Cheng, Li},
  booktitle={2024 IEEE/CVF Conference on Computer Vision and Pattern Recognition (CVPR)},
  pages={1900--1910},
  year={2024},
  organization={IEEE}
}

@inproceedings{li2026motionhiflow,
  title     = {MotionHiFlow: Text-to-Motion via Hierarchical Flow Matching},
  author    = {Li, Heng and Lin, Xiaotong and Zeng, Ling-An and Kang, Yulei and Li, Shuai and Hu, Jian-Fang},
  booktitle = {Proceedings of the IEEE/CVF Conference on Computer Vision and Pattern Recognition (CVPR)},
  year      = {2026}
}

@inproceedings{tevet2022motionclip,
  title={Motionclip: Exposing human motion generation to clip space},
  author={Tevet, Guy and Gordon, Brian and Hertz, Amir and Bermano, Amit H and Cohen-Or, Daniel},
  booktitle={European conference on computer vision},
  pages={358--374},
  year={2022},
  organization={Springer}
}

@inproceedings{li2025lamp,
  title={Lamp: Language-motion pretraining for motion generation, retrieval, and captioning},
  author={Li, Zhe and Yuan, Weihao and He, Yisheng and Qiu, Lingteng and Zhu, Shenhao and Gu, Xiaodong and Shen, Weichao and Dong, Yuan and Dong, Zilong and Yang, Laurence},
  booktitle={International Conference on Learning Representations},
  volume={2025},
  pages={84238--84250},
  year={2025}
}

@article{radford2021clip,
  title={Learning transferable visual models from natural language supervision},
  author={Radford, A},
  journal={arXiv preprint arXiv:2103.00020},
  year={2021}
}

@InProceedings{sun2026amr,
    author    = {Sun, Shengkai and Cheng, Zhiyong and Zhang, Zefan and Dong, Jianfeng and Li, Zhihui and Wang, Meng},
    title     = {Exploring Adaptive Masked Reconstruction for Self-Supervised Skeleton-Based Action Recognition},
    booktitle = {Proceedings of the IEEE/CVF Conference on Computer Vision and Pattern Recognition (CVPR)},
    month     = {June},
    year      = {2026},
    pages     = {13974-13983}
}

@inproceedings{jang2020etri,
  title={ETRI-activity3D: A large-scale RGB-D dataset for robots to recognize daily activities of the elderly},
  author={Jang, Jinhyeok and Kim, Dohyung and Park, Cheonshu and Jang, Minsu and Lee, Jaeyeon and Kim, Jaehong},
  booktitle={2020 IEEE/RSJ International Conference on Intelligent Robots and Systems (IROS)},
  pages={10990--10997},
  year={2020},
  organization={IEEE}
}

@inproceedings{wang2014nwucla,
  title={Cross-view action modeling, learning and recognition},
  author={Wang, Jiang and Nie, Xiaohan and Xia, Yin and Wu, Ying and Zhu, Song-Chun},
  booktitle={Proceedings of the IEEE conference on computer vision and pattern recognition},
  pages={2649--2656},
  year={2014}
}

@inproceedings{li2010msraction3d,
  title={Action recognition based on a bag of 3d points},
  author={Li, Wanqing and Zhang, Zhengyou and Liu, Zicheng},
  booktitle={2010 IEEE computer society conference on computer vision and pattern recognition-workshops},
  pages={9--14},
  year={2010},
  organization={IEEE}
}

@inproceedings{xia2012utkinect,
  title={View invariant human action recognition using histograms of 3d joints},
  author={Xia, Lu and Chen, Chia-Chih and Aggarwal, Jake K},
  booktitle={2012 IEEE computer society conference on computer vision and pattern recognition workshops},
  pages={20--27},
  year={2012},
  organization={IEEE}
}

@inproceedings{seidenari2013florence,
  title={Recognizing actions from depth cameras as weakly aligned multi-part bag-of-poses},
  author={Seidenari, Lorenzo and Varano, Vincenzo and Berretti, Stefano and Bimbo, Alberto and Pala, Pietro},
  booktitle={Proceedings of the IEEE conference on computer vision and pattern recognition workshops},
  pages={479--485},
  year={2013}
}

@inproceedings{yun2012sbu,
  title={Two-person interaction detection using body-pose features and multiple instance learning},
  author={Yun, Kiwon and Honorio, Jean and Chattopadhyay, Debaleena and Berg, Tamara L and Samaras, Dimitris},
  booktitle={2012 IEEE computer society conference on computer vision and pattern recognition workshops},
  pages={28--35},
  year={2012},
  organization={IEEE}
}

@inproceedings{wu2024scdnet,
  title={Scd-net: Spatiotemporal clues disentanglement network for self-supervised skeleton-based action recognition},
  author={Wu, Cong and Wu, Xiao-Jun and Kittler, Josef and Xu, Tianyang and Ahmed, Sara and Awais, Muhammad and Feng, Zhenhua},
  booktitle={Proceedings of the AAAI conference on artificial intelligence},
  volume={38},
  number={6},
  pages={5949--5957},
  year={2024}
}

@inproceedings{wu2024macdiff,
  title={Macdiff: Unified skeleton modeling with masked conditional diffusion},
  author={Wu, Lehong and Lin, Lilang and Zhang, Jiahang and Ma, Yiyang and Liu, Jiaying},
  booktitle={European conference on computer vision},
  pages={110--128},
  year={2024},
  organization={Springer}
}

@inproceedings{do2024skateformer,
  title={Skateformer: skeletal-temporal transformer for human action recognition},
  author={Do, Jeonghyeok and Kim, Munchurl},
  booktitle={European Conference on Computer Vision},
  pages={401--420},
  year={2024},
  organization={Springer}
}

@inproceedings{sun2025gfp,
  title={Towards efficient general feature prediction in masked skeleton modeling},
  author={Sun, Shengkai and Zhang, Zefan and Dong, Jianfeng and Cheng, Zhiyong and Chang, Xiaojun and Wang, Meng},
  booktitle={2025 IEEE/CVF International Conference on Computer Vision (ICCV)},
  pages={12212--12221},
  year={2025},
  organization={IEEE}
}

@inproceedings{mehraban2026stars,
  title={STARS: Self-supervised tuning for 3D action recognition in skeleton sequences},
  author={Mehraban, Soroush and Rajabi, Mohammad Javad and Iaboni, Andrea and Taati, Babak},
  booktitle={2026 IEEE/CVF Winter Conference on Applications of Computer Vision (WACV)},
  pages={2858--2868},
  year={2026},
  organization={IEEE}
}

@article{gunasekara2026dimop,
  title={DiMoP: Diffusion-Driven Motion Representation Learning With Frame-Level Pseudo-Classification for Skeleton-Based Action Recognition},
  author={Gunasekara, Shanaka Ramesh and Li, Wanqing and Kaldera, Nikalal and Ogunbona, Philip and Yang, Jack},
  journal={IEEE Transactions on Biometrics, Behavior, and Identity Science},
  year={2026},
  publisher={IEEE}
}

@inproceedings{singh2026pathjepa,
  title={Path-JEPA: Path Signature Based Predictive Learning for Skeleton Action Recognition},
  author={Singh, Ashutosh and Singh, Ashish},
  booktitle={European Conference on Computer Vision},
  pages={110--128},
  year={2026},
  organization={Springer}
}

@inproceedings{wu2023skeletonmae,
  title={Skeletonmae: Spatial-temporal masked autoencoders for self-supervised skeleton action recognition},
  author={Wu, Wenhan and Hua, Yilei and Zheng, Ce and Wu, Shiqian and Chen, Chen and Lu, Aidong},
  booktitle={2023 IEEE international conference on multimedia and expo workshops (ICMEW)},
  pages={224--229},
  year={2023},
  organization={IEEE}
}

@inproceedings{tran2013sparse,
  title={Sparse spatio-temporal representation of joint shape-motion cues for human action recognition in depth sequences},
  author={Tran, Quang D and Ly, Ngoc Q},
  booktitle={The 2013 RIVF International Conference on Computing \& Communication Technologies-Research, Innovation, and Vision for Future (RIVF)},
  pages={253--258},
  year={2013},
  organization={IEEE}
}

@inproceedings{wang2016snippets,
  title={Recognizing actions in 3d using action-snippets and activated simplices},
  author={Wang, Chunyu and Flynn, John and Wang, Yizhou and Yuille, Alan},
  booktitle={Proceedings of the AAAI Conference on Artificial Intelligence},
  volume={30},
  number={1},
  year={2016}
}

@article{luvizon2017harskel,
  title={Learning features combination for human action recognition from skeleton sequences},
  author={Luvizon, Diogo Carbonera and Tabia, Hedi and Picard, David},
  journal={Pattern Recognition Letters},
  volume={99},
  pages={13--20},
  year={2017},
  publisher={Elsevier}
}

@inproceedings{wen2023istanet,
  title={Interactive spatiotemporal token attention network for skeleton-based general interactive action recognition},
  author={Wen, Yuhang and Tang, Zixuan and Pang, Yunsheng and Ding, Beichen and Liu, Mengyuan},
  booktitle={2023 IEEE/RSJ International Conference on Intelligent Robots and Systems (IROS)},
  pages={7886--7892},
  year={2023},
  organization={IEEE}
}

@inproceedings{liu2025protogcn,
  title={Revealing key details to see differences: A novel prototypical perspective for skeleton-based action recognition},
  author={Liu, Hongda and Liu, Yunfan and Ren, Min and Wang, Hao and Wang, Yunlong and Sun, Zhenan},
  booktitle={2025 IEEE/CVF Conference on Computer Vision and Pattern Recognition (CVPR)},
  pages={29248--29257},
  year={2025},
  organization={IEEE}
}

@inproceedings{zhou2025hypergcn,
  title={Adaptive hyper-graph convolution network for skeleton-based human action recognition with virtual connections},
  author={Zhou, Youwei and Xu, Tianyang and Wu, Cong and Wu, Xiaojun and Kittler, Josef},
  booktitle={2025 IEEE/CVF International Conference on Computer Vision (ICCV)},
  pages={12648--12658},
  year={2025},
  organization={IEEE}
}

@inproceedings{zhou2026gamba,
  title={Gamba: Mamba-based graph convolutional network with dynamic graph topology learning for action recognition},
  author={Zhou, Rouyi and Wu, Yangzhi and Wen, Jiajun and Gao, Can and Liu, Feng and Lai, Zhihui and Shen, Linlin},
  booktitle={Proceedings of the IEEE/CVF Conference on Computer Vision and Pattern Recognition},
  pages={6065--6074},
  year={2026}
}

@inproceedings{li2026jit,
  title={Back to basics: Let denoising generative models denoise},
  author={Li, Tianhong and He, Kaiming},
  booktitle={Proceedings of the IEEE/CVF Conference on Computer Vision and Pattern Recognition},
  pages={36115--36125},
  year={2026}
}

@inproceedings{yu2026pixeldit,
  title={Pixeldit: Pixel diffusion transformers for image generation},
  author={Yu, Yongsheng and Xiong, Wei and Nie, Weili and Sheng, Yichen and Liu, Shiqiu and Luo, Jiebo},
  booktitle={Proceedings of the IEEE/CVF Conference on Computer Vision and Pattern Recognition},
  pages={14273--14282},
  year={2026}
}

@misc{loshchilov2019adamw,
      title={Decoupled Weight Decay Regularization}, 
      author={Ilya Loshchilov and Frank Hutter},
      year={2019},
      eprint={1711.05101},
      archivePrefix={arXiv},
      primaryClass={cs.LG},
      url={https://arxiv.org/abs/1711.05101}, 
}

@inproceedings{duan2022revisiting,
  title={Revisiting skeleton-based action recognition},
  author={Duan, Haodong and Zhao, Yue and Chen, Kai and Lin, Dahua and Dai, Bo},
  booktitle={2022 IEEE/CVF Conference on Computer Vision and Pattern Recognition (CVPR)},
  pages={2959--2968},
  year={2022},
  organization={IEEE}
}

@inproceedings{abdelfattah2024maskclr,
  title={Maskclr: Attention-guided contrastive learning for robust action representation learning},
  author={Abdelfattah, Mohamed and Hassan, Mariam and Alahi, Alexandre},
  booktitle={2024 IEEE/CVF Conference on Computer Vision and Pattern Recognition (CVPR)},
  pages={18678--18687},
  year={2024},
  organization={IEEE}
}

@inproceedings{zhou2024blockgcn,
  title={Blockgcn: Redefine topology awareness for skeleton-based action recognition},
  author={Zhou, Yuxuan and Yan, Xudong and Cheng, Zhi-Qi and Yan, Yan and Dai, Qi and Hua, Xian-Sheng},
  booktitle={2024 IEEE/CVF Conference on Computer Vision and Pattern Recognition (CVPR)},
  pages={2049--2058},
  year={2024},
  organization={IEEE}
}

@article{peng2026sbf,
  title={SBF: An Effective Representation to Augment Skeleton for Video-based Human Action Recognition},
  author={Peng, Zhuoxuan and Ding, Yiyi and Lin, Yang and Chan, S-H Gary},
  journal={arXiv preprint arXiv:2604.03590},
  year={2026}
}

@inproceedings{yan2023skeletonmae,
  title={Skeletonmae: graph-based masked autoencoder for skeleton sequence pre-training},
  author={Yan, Hong and Liu, Yang and Wei, Yushen and Li, Zhen and Li, Guanbin and Lin, Liang},
  booktitle={2023 IEEE/CVF International Conference on Computer Vision (ICCV)},
  pages={5583--5595},
  year={2023},
  organization={IEEE}
}

@inproceedings{sun2019hrnet,
  title={Deep high-resolution representation learning for human pose estimation},
  author={Sun, Ke and Xiao, Bin and Liu, Dong and Wang, Jingdong},
  booktitle={2019 IEEE/CVF conference on computer vision and pattern recognition (CVPR)},
  pages={5686--5696},
  year={2019},
  organization={IEEE}
}

@article{fischler1981random,
  title={Random sample consensus: a paradigm for model fitting with applications to image analysis and automated cartography},
  author={Fischler, Martin A and Bolles, Robert C},
  journal={Communications of the ACM},
  volume={24},
  number={6},
  pages={381--395},
  year={1981},
  publisher={ACM, New York, NY, USA}
}

@inproceedings{guo2020action2motion,
  title={Action2motion: Conditioned generation of 3d human motions},
  author={Guo, Chuan and Zuo, Xinxin and Wang, Sen and Zou, Shihao and Sun, Qingyao and Deng, Annan and Gong, Minglun and Cheng, Li},
  booktitle={Proceedings of the 28th ACM international conference on multimedia},
  pages={2021--2029},
  year={2020}
}
}

\end{document}